%% file: arXiv.tex
\documentclass[10pt,onecolumn,letterpaper]{article}

\usepackage[top=1.5in,bottom=1.5in,left=1.2in,right=1.2in,columnsep=0.35in]{geometry}

\input{macro&setup/myPKG}

\input{macro&setup/myDeco}

\input{macro&setup/myMacro}

\usepackage{indentfirst}
\usepackage{xcolor}
\usepackage{hyperref}
\usepackage{enumitem}
\usepackage{setspace}
\usepackage{algorithm}
\usepackage{algorithmic}
\usepackage{amsmath}
\usepackage{amsthm}
\usepackage{amssymb}
\usepackage{natbib}
\usepackage{stmaryrd}
\usepackage{hyperref}
\usepackage{tabularx}
\usepackage{booktabs} 
\usepackage{color}
\usepackage{graphicx}
\usepackage{caption}

\usepackage{authblk}

\author[1]{Chi-Hua Wang}
\author[2]{Wenjie Li}

\affil[1]{   University of California, 
       Los Angeles}
\affil[2]{Department of Statistics, Purdue University }
\date{}

\begin{document}

\title{Always-Valid Risk Bounds for Low-Rank Online Matrix Completion}


\maketitle

\begin{abstract}
Always-valid concentration inequalities are increasingly used as performance measures for online statistical learning, notably in the learning of generative models and supervised learning. Such inequality advances the online learning algorithms design by allowing random, adaptively chosen sample sizes instead of a fixed pre-specified size in offline statistical learning. However, establishing such an always-valid type result for the task of matrix completion is challenging and far from understood in the literature. Due to the importance of such type of result, this work establishes and devises the always-valid risk bound process for online matrix completion problems. Such theoretical advances are made possible by a novel combination of non-asymptotic martingale concentration and regularized low-rank matrix regression. Our result enables a more sample-efficient online algorithm design and serves as a foundation to evaluate online experiment policies on the task of online matrix completion.
\end{abstract}

\section{Introduction}
\label{sec: intro}


We consider an online matrix completion problem with continuous monitoring, a setting where decision-makers seek to recover a structured matrix from noisy partial measurements (matrix completion), while the decision-makers are allowed to terminate the experiment whenever they wish, \textit{and} the result still maintains statistical validity (continuous monitoring). Such a setting arises naturally in industrial practice but remains challenging in the literature, preventing practitioners from effectively deploying matrix completion methodology  in modern online service industries.

Risk control of learned models in continuously-monitored online experiments is in emerging demand from industrial practice because the opportunity cost of lengthy experiments is high and regrettable (\cite{johari2021always}).
Indeed, it is desirable to detect the true effect size as quickly as possible, or to abolish the running experiment if the effect appears unpromising so that scientists may test other available actions. Besides, optimizing the running time in advance is unfortunately impractical due to the lack in prior knowledge on the seeking effect size and cost elasticity. In modern online experiment practice, the deployment of online statistical learning methodology turns out to be impeded by such dynamic trade-off between maximum effect detection and minimum running time. Resolving such dynamic trade-off is a crucial advancement of real-time data learning methodology, which is pioneered by \cite{wang2020online} in the dynamic pricing setting, but it still remains an open question in the setting of online matrix completion problems.

\begin{wrapfigure}{r}{7cm}
\centering
\includegraphics[width=\linewidth]{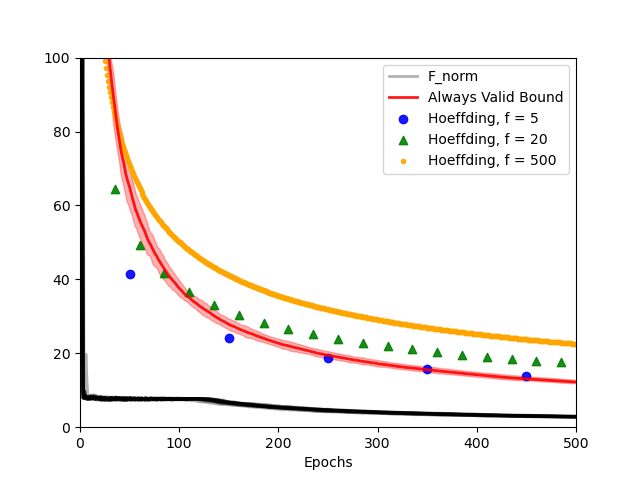}
\caption{\textit{Continuous monitoring by our proposed always-valid bound versus Sequential monitoring by Hoeffding bound given pre-specified evaluation at 5, 20, 500 time points. In sequential monitoring method, the decision makers needs to pre-specify a set of ''check points'' and pre-allocate confidence budget for each checkpoint. 
In contrast, continuous monitoring method allows decision makers to check the result anytime they want with single confidence budget. Such anytime validity is a great benefit on developing the performance monitor in applications of online statistical learning. }
}
\label{fig:AlwaysValidBound}
\end{wrapfigure}
Risk bounds for the estimation error in the offline matrix completion problem (\cite{Negahban2012A, lafond2015online, wainwright2019high}) are the main analytic tools to evaluate and control the online statistical learning performance and risk. These bounds are used to limit the probability of a large discrepancy between the estimation and the ground-truth parameter matrix for a \textit{pre-specified fixed} number of samples $n$. The fundamental rationality of such risk control is to utilize the unlikely occurrence of large deviations to support high-confidence conclusions for online decision-making. Yet, these bounds only hold for a pre-specified fixed constant number of samples, which could lead to suboptimal algorithms. Consequently, a more efficient online matrix completion algorithm design demands the ability to draw always-valid conclusions even when the number of samples itself is a random variable. Such lack of validity in randomly-stopped online matrix completion algorithm inspires our investigation on the continuous monitoring online matrix completion problem.



Solving this fundamental problem based on existing knowledge of offline matrix completion (\cite{Negahban2012A, lafond2015online, wainwright2019high}) requires three major rethinkings: 
\textbf{(a) Rethinking task modeling from offline to online.} Depart from the i.i.d. noise assumptions in offline matrix completion literature, our framework should allow dependent noise process in modeling the task of matrix completion to better captured the nature in online service industries. 
\textbf{(b) Rethinking regularization from static dataset to dynamic datastream.} In offline statistical learning,
the trade-off between \textit{static dataset} fitting and structural privileging has been designed and fine-tuned with a \textit{fixed} regularization level; however, whether such fixed regularization is appropriate to trade off \textit{dynamic datastream} fitting demands rethinking on the essence of regularization.  
\textbf{(c) Rethinking performance monitoring from pre-defined time to any time.} In sequential monitoring, the task learning performance is sequentially evaluated online at a sequence of \textit{pre-defined} time, raising concerns of the statistical validity if we checked the task performance outside the pre-defined time points; how to maintain statistical validity from pre-defined time to anytime demands rethinking the essence of online uncertainty.

\begin{table*}[h!]
\centering
    \footnotesize
   \vspace{5pt}
\begin{tabular}{l l}
\hline
\vspace{2pt}
{Notation} & {Description}  \\
\hline
\vspace{5pt}
$\vertiii{\mathbf{X}}_F$ & The Frobenius norm, i.e., $\vertiii{\mathbf{X}}_F = (\sum_{i=1}^{d_1} \sum_{j=1}^{d_2} \mathbf{X}_{ij}^2)^{1/2}$. \\
\vspace{5pt}
$\vertiii{\mathbf{X}}_{nuc}$ & The nuclear norm, i.e., $\vertiii{\mathbf{X}}_{nuc} = \sum_{i=1}^{\min \{d_1, d_2\} } \sigma_i$, where $\{\sigma_i\}$'s are the singular values of $\mathbf{X}$. \\
\vspace{5pt}
$\vertiii{\mathbf{X}}_{op}$ & The operator norm induced by the vector norm $\|\cdot\|_2$, i.e., $\vertiii{\mathbf{X}}_{op} = \sup_{n\in \mathbb{R}^d} \|\mathbf{X}n\|_2/ \|n\|_2$.  \\
\vspace{5pt}
$\vertiii{\mathbf{X}}_{max}$ & The element-wise maximum norm, i.e., $\vertiii{\mathbf{X}}_{max} = \max_{i=1}^{d_1} \max_{j=1}^{d_2} |\mathbf{X}_{ij}|$. \\
\hline
\end{tabular}
\caption{\footnotesize Table of matrix norms used throughout this paper. $\mathbf{X} \in \mathbb{R}^{\rowdim \times \coldim}$ is a matrix }
\label{tab: matrix_norms}
\end{table*}

\subsection{Main contributions}

We summarize the knowledge advancement of this work
contributing to existing matrix completion methods 
(that learn a static model with a fixed regularization level and have validity at some pre-specified sample size):

\textbf{(i) The first online matrix completion procedure that allows continuous monitoring of the prediction risk.} We deliver the first algorithm design framework on the construction of online regularization sequence for continuous monitoring online matrix completion problems. Our framework improves the existing offline matrix completion efforts, e.g., \citet{lafond2015low} by allowing dependent noise process and prediction error continuous monitoring. In particular, our design of ''online regularization sequence'' advances the prior art in \citet{wang2020online} by allowing  non-parametric noise class in matrix completion models. Our framework provides  precise online regularization sequence for both  sub-Gaussian noise class (Lemma \ref{lem: choice_of_lambda_subgaussian}) and sub-Exponential noise class (Lemma \ref{lem: choice_of_lambda_subexponential}), leading to a novel always-valid prediction risk bound process (Theorem \ref{thm: oracle_inequality}) to meet the demand of continuously monitored online statistical learning in modern online service industrial practice.  With the continuous monitoring framework, the decision makers can check the result anytime they want with single confidence budget. Such anytime validity removes the confidence pre-allocation issue in sequential monitoring methods and thus is a great benefit on developing the performance monitor in applications of online statistical learning.

\textbf{(ii) A theoretical framework of continuous monitoring online statistical learning.} Section \ref{sec:theory_results} delivers a three-stage general framework towards continuous monitoring of online statistical learning: regularization sequence design, gradient norm control and alway-valid prediction error process. 
We establish the first always-valid risk bound results for continuous monitoring in the online matrix completion problem.
The key to achieving such always valid guarantee is to design an ''online regularization sequence'' to trade off dynamically between accuracy and uncertainty in noisy matrix completion.  Such theoretical advances are made possible by our novel use of non-asymptotic martingale concentration to impose optimistic uncertainty control on the online statistical learning process. As a technical byproduct, we develop non-asymptotic matrix martingale concentration for self-adjoint matrix martingale difference process arise from general non-parametric class including sub-Gaussian and sub-Exponential noise process.

\textbf{(iii) Empirical results comparing sequential monitoring and continuous monitoring on prediction risk process.} Section \ref{sec: experiments} delivers empirical results on comparing sequential monitoring, which are subject to pre-specified checking points, with our novel use-case: continuous monitoring. Our always-valid prediction risk bound enable decision maker to check the progress of online matrix completion anytime they want with uniform statistical validity. We show the established always valid prediction bound process successfully perform continuous monitoring across different models including Gaussian, Binomial and Poisson online matrix completion. Our results support the benefits on anytime validity of continuous monitoring framework over sequential monitoring via Hoeffiding bound.

%
%







\subsection{Related Works}

\textbf{Offline noisy matrix completion.} Risk bounds for fixed sample-size noisy matrix completion problems had been systematically studied in the last decade ( \cite{gunasekar2014exponential, klopp2014noisy, lafond2015low, negahban2012restricted, elsener2018robust}). While these efforts inspired several interesting concepts of proof and elegant theoretical frameworks under the topic of high-dimensional statistics (\cite{ Negahban2012A,wainwright2019high}), they failed to meet the challenge from online statistical learning with continuous monitoring (\cite{ johari2021always, zhao2016adaptive}) demanded in modern online service industry. Our work brides the gap between offline and online noisy matrix completion.  

\textbf{Sequential and continuous monitoring.}
Sequential monitoring methodology is a standard practice in modern covariate-adaptive randomized clinical trials although the validity of this sequential procedure is not well studied in the literature \citep{zhu2019sequential}. Recently, sequential monitoring has been adopted to dynamic causal effects evaluation in A/B testing \citep{shi2022dynamic}. Removing the confidence allocation issue in sequential monitoring, continuous monitoring methodology \citep{johari2021always} serves as the theoretical foundation of modern online experiment methodology. Our work contributes to continuous monitoring literature by constructing the prediction risk monitors of online matrix completion. 

\textbf{Non-asymptotic martingale concentration.}
Advancing the prior art on always-valid lasso oracle inequality pioneered in \cite{wang2020online}, we utilize the current art of non-asymptotic matrix concentration (\cite{koltchinskii2013remark, Tropp2011User, Tropp2011freedmans, howard2021time, maillard2019mathematics}) and successfully establish the first always-valid risk bound process for online matrix completion problem.
 
\subsection{Notations}
For any positive integers $d_1, d_2, n \in \mathbb{N}$, denote $d_{1} \vee d_{2}:=\max \left\{d_{1}, d_{2}\right\}$ , $d_{1} \wedge d_{2}:=\min \left \{d_{1}, d_{2}\right \}$,  and $[n]:=\{1,2,3, \cdots, n\}$. We equip the matrix space $\mathbb{R}^{\rowdim \times \coldim}$ with the \textit{trace inner product $\TrInnProd{\cdot}{ \cdot}$} (Hilbert-Schmidt inner product), i.e., for any two matrices $\mathbf{A}, \mathbf{B} \in \mathbb{R}^{\rowdim \times \coldim}$, their inner product is denoted by
\mydeco{equation}{
\label{eq:TraceInnProd}
\TrInnProd{\mathbf{A}}{ \mathbf{B}}
:=\operatorname{trace}\left(\mathbf{A}^{\mathrm{T}} \mathbf{B}\right)=\sum_{i=1}^{\rowdim} \sum_{j=1}^{\coldim} \textbf{A}_{i j} \mathbf{B}_{i j}
}
We use $\vertiii{\cdot}$ to denote different matrix norms. A complete list of the norms used in this paper with their definitions is provided in Table \ref{tab: matrix_norms}. For any differentiable function $\Phi:\mathbb{R}^d \mapsto \mathbb{R}$, $\nabla \Phi$ denotes its gradient. For any $x, y \in \mathbb{R}^d$, the Bregman divergence between $x$ and $y$ induced by $\Phi$ is defined by 
\begin{equation}
\label{eq:Breg_div}
    B_{\Phi}(x,y) = \Phi(x) - \Phi(y) - \langle \nabla \Phi(y), x-y \rangle
\end{equation}
We use $\widetilde{\mathcal{O}}(\cdot)$ to hide poly-logarithmic factors in big-$\mathcal{O}$ notations.







\section{Problem formulation}

This section elaborates the problem formulation for online noisy matrix completion. 
Section \ref{subsec:AV_prediction} defines the concept of always-valid prediction risk bound process. Section \ref{subsec:online_matrix_completion} formulates the online matrix completion problem. Section \ref{subsec:stat_reg_cond} provides essential statistical regularity conditions for the main results in Section \ref{sec:theory_results}.

\textbf{Model specifications.} Our aim is to recover an unknown low-rank target matrix $\mathbf{\Theta}^{*} \in \mathbb{R}^{d_1 \times d_2}$ with rank $r \leq d_1 \wedge d_2 $.  A sampling policy $\Pi$ is defined as a sequence of indexes of length $T$,
%
%
$\left\{\pi_{t}\right\}_{t=1}^{T} \in\left(\left[d_{1}\right] \times\left[d_{2}\right]\right)^{T}$, that observes noisy individual entries of $\mathbf{\Theta}^{*}$. Formally, let $\mathbf{E}_{\pi_i} = e_{k_t} e_{l_t}^T$, where $\{e_{k}\}_{k=1}^{\rowdim}$ and $\{e_{l}\}_{l=1}^{\coldim}$ are the standard basis of $\mathbb{R}^{\rowdim}$ and $\mathbb{R}^{\coldim}$. Then the policy $\Pi$ observes a noisy observation of  element at time $t$
\begin{equation}
  \Signal{t} = \TrInnProd{\mathbf{E}_{\pi_t}}{\mathbf{\Theta}^*}.
\end{equation}
The noisy observation $y_{t}$ is modeled by the mechanism 
\begin{equation}
\label{eq:regression_model}
   \Obs_t = g\big( \Signal{t}\big) + \epsilon_t.
\end{equation}
The function $g(\cdot)$ is often termed as the link function. In practice, the function $g(\cdot)$ is user-specified and it is designed to adapt to the application domain. For example, for the choices of $G(\cdot)$ in Table \ref{table:log_partition}, by setting $g(\cdot) = G^{'}(\cdot)$ in model \eqref{eq:regression_model}, we recover different exponential family models (See Remark \ref{rm:ExpFamModel} for details).

\mydeco{remark}{The model class  \eqref{eq:regression_model} generalizes the exponential family models considered in the matrix completion literature \citep{lafond2015low}. \label{rm:ExpFamModel} Given the filtration $\mathcal{H}_t$ at time $t$, the observation $y_t$ is assumed to follow a natural exponential family distribution, conditionally on the selected $\TruePara$ entry; that is, the conditional distribution of $\Obs_{t} \mid \Signal{t}$ is given by $\operatorname{Exp}_{h, G}
\left(\Signal{t}\right)$, where
\mydeco{equation}{
\nonumber
\label{eq: condit_model}
\begin{aligned}
&
\operatorname{Exp}_{h, G}
\left(\Signal{t}\right)
:=h\left(\Obs_{t}\right)\exp(\Signal{t} \Obs_{t}-G\big(\Signal{t}\big)).
\end{aligned}
} 
Associated to the canonical representation, 
$h$ is the base measure and $G: \mathbb{R} \mapsto \mathbb{R}$, is a strictly convex, analytic function called the log-partition function. Examples of different canonical representations and their log-partition functions are listed in Table \ref{table:log_partition}.
 }

\textbf{Summary statistics for policy sequence.} 
At each iteration $t$, we define $p_t(k,l)$ to be the empirical frequency of the index $(k,l)$ in all the samples generated by the policy $\Pi$, i.e. $p_t(k,l) =  \frac{1}{t} \sum_{i=1}^t \mathbbm{1} \{\pi_i = (k,l)\}$.
We then define the term $\bar{p}_{t}$ to denote the minimal empirical sampling frequency at time $t$; formally, 
\begin{equation}
\label{eq:min_samp_frequency}
\bar{p}_t = \min_{k\in[p_1], l \in [p_2]} p_t(k,l)
\end{equation}
We  also define the following term $S_t$ at each iteration $t$ that measures the variation of the selection policy.
\begin{equation}
\label{eq: Sigma}
{S}_t = \max \left \{\vertiii{\frac{1}{t}\sum_{i=1}^t \mathbf{E}_{\pi_i}  \mathbf{E}_{\pi_i}^T}_{op}, \vertiii{\frac{1}{t} \sum_{i=1}^t \mathbf{E}_{\pi_i}^T \mathbf{E}_{\pi_i}}_{op}\right \}
\end{equation}
The above definition is to emphasize the dependency on the policy $\pi$. In the context of matrix completion, the equation \eqref{eq: Sigma} is equivalent to 
\begin{equation*}
{S}_t = \max \left \{\vertiii{\frac{1}{t}\sum_{i=1}^t 
e_{k_i} e_{k_j}^\top
}_{op}, \vertiii{\frac{1}{t} 
\sum_{i=1}^t e_{k_j} e_{k_i}^\top
}_{op}\right \}
\end{equation*}

\begin{remark} Indeed, how to design a 'good' online experiment policy $\Pi$ is an interesting and important problem that goes beyond the scope of our current study, where we assume the policy is given and the task is to measure the task performance of online matrix completion. Our always valid risk bound process can serve as a choice of performance measure. How to do policy improvement based on our always valid risk bound process towards more sample efficient online matrix completion is an exciting future direction.
\end{remark}

\begin{table}[t!]
\centering
\begin{tabular}{|l|c|c|}
\hline Distribution & Parameter $x$ & $G(x)$ \\
\hline Gaussian: $\mathcal{N}\left(\mu, \sigma^{2}\right)$ & $\mu / \sigma$ & $\sigma^{2} x^{2} / 2$ \\
\hline Binomial: $\mathcal{B}^{N}(p)$ & $\log (\frac{p}{1-p})$ & $N \log \left(1+\mathrm{e}^{x}\right)$ \\
\hline Poisson: $\mathcal{P}(\lambda)$ & $\log (\lambda)$ & $\mathrm{e}^{x}$ \\
\hline Exponential: $\mathcal{E}(\lambda)$ & $-\lambda$ & $-\log (-x)$ \\
\hline
\end{tabular}
\caption{Canonical representations and their log-partition functions \citep{lafond2015low}}
\label{table:log_partition}
\end{table}


\subsection{Always-Valid Prediction} 
\label{subsec:AV_prediction}

Our main objective is to construct the following \textit{always-valid prediction risk bound process}:

\mydeco{definition}{\label{def:AVRB} 
Given any (possibly unbounded) stopping time $T$, 
a sequence of constant real numbers $\{r_{t}\}_{t=1}^{T}$ is an {always valid prediction risk bound sequence} of the estimator sequence $\widehat{\mathbf{\Theta}}_{1}, \widehat{\mathbf{\Theta}}_{2}, \cdots, \widehat{\mathbf{\Theta}}_{T}$ with confidence budget $\alpha$ if  
\begin{equation}
\label{eq:EP_AlwaysValid_RiskProcess}
    \mathbb{P}\bigg( \big\{ \exists t \leq T  :  \vertiii{\widehat{\mathbf{\Theta}}_{t} - \mathbf{\Theta}^*}_F \ge r_{t}\big\}\bigg) \leq \alpha
\end{equation}
}

Definition \ref{def:AVRB} aims to allow decision-makers to terminate the online matrix completion task whenever they wish \textit{but} still have statistical guarantees. Such a goal is important because it enables the decision-makers to trade off the prediction risk of the online-learned model with run-time (sampling budget) under a user-specified confidence budget.


\subsection{Online matrix completion}
\label{subsec:online_matrix_completion}

\textbf{Online matrix completion procedure.} Our matrix completion estimator is defined as the minimizer of the nuclear-norm penalized self-information loss function.
Specifically, at a time step $t$, for a target matrix $\mathbf{\Theta} \in \mathbb{R}^{d_1\times d_2}$, we denote $\Phi_t(\mathbf{\Theta})$ to be the negative log-likelihood loss, defined as:  
\begin{equation}
\label{eq:work_loss}
    \begin{aligned}
       \Phi_t(\mathbf{\Theta}) =  \frac{1}{t} \sum_{i=1}^t \left [G\left(\mathbf{\Theta}_{\pi_i})\right)- \mathbf{\Theta}_{\pi_i} \Obs_i  \right]
    \end{aligned}
\end{equation}
For any constant $\gamma >0$ and any regularization level $\lambda_{t} >0$, the nuclear norm penalized estimator $\widehat{\mathbf{\Theta}}_{t}$ is defined as:
\begin{equation}
\label{eq:work_program}
    \begin{aligned}
    \widehat{\mathbf{\Theta}}_{t} \equiv \arg\min_{\vertiii{\mathbf{\Theta}}_{\max}\le \gamma}
        \Phi_t(\mathbf{\Theta})+ \lambda_t \vertiii{\mathbf{\Theta}}_{nuc}
    \end{aligned}
\end{equation}
Repeating the above matrix completion procedure at each time step $t = 1,2,\cdots, T$ produces a sequence of online-learned matrix completion estimators $\{\widehat{\mathbf{\Theta}}_{1}, \widehat{\mathbf{\Theta}}_{2}, \cdots, \widehat{\mathbf{\Theta}}_{T}\}$. We are interested in providing statistical guarantees on the prediction error sequence:
\begin{equation}
\label{eq:predic_err_process}
\left \{\vertiii{\mathbf{\widehat{\Theta}}_1 - \mathbf{\Theta^*}}_{F}, \vertiii{\mathbf{\widehat{\Theta}}_2 - \mathbf{\Theta^*}}_{F}, \cdots, \vertiii{\mathbf{\widehat{\Theta}}_T - \mathbf{\Theta^*}}_{F}
\right\}
\end{equation}
%

The fundamental challenge of analyzing \eqref{eq:predic_err_process} is to choose an \textit{online regularization sequence} 
\begin{equation}
\label{eq:Ref_Process}
\lambda_{1}, \lambda_{2}, \cdots, \lambda_{T}
\end{equation}
at each time of matrix completion procedure \eqref{eq:work_program}. Intuitively, the regularization level $\lambda_{t}$ controls the trade off between fitting the data and privileging a low rank solution: for large $\lambda_{t}$, the rank of $\widehat{\mathbf{\Theta}}_{t}$ is expected to be small. How to construct an online regularization sequence \ref{eq:Ref_Process} and construct the corresponding always-valid prediction process (Definition \ref{def:AVRB}) is the main technical challenge of this paper. 


\begin{remark} (The reason why high-dimensional statistics techniques are not directly applicable to analyzing \eqref{eq:predic_err_process}.) Note that the techniques developed in the literature of high-dimensional statistics for matrix completion problem \eqref{eq:work_program}(e.g. \cite{wainwright2019high, Negahban2012A}) only works for a \textbf{fixed} time horizon $T$. Such techniques, unfortunately, are not directly applicable to our case, where the time horizon $T$ is allowed to be a \textit{random} time. 

\end{remark}

\subsection{Regularity conditions}
\label{subsec:stat_reg_cond}

Here we provide the statistical regularity assumptions for the analysis of the matrix completion procedure in \eqref{eq:work_program}, which are all standard assumptions in the matrix completion problem.

\begin{assumption}
\label{assumption: bound_of_Theta}
The target matrix $\mathbf{\Theta}^*$ has bounded elements, i.e., there exists a constant $\gamma >0$ s.t.
\begin{equation}
\vertiii{\mathbf{\Theta}^*}_{max} \leq \gamma
\end{equation} 
\end{assumption}
For example, in a recommendation system, the constant $\gamma$ is the highest rank of a product.

\begin{assumption}
\label{assumption: bounds_of_ddG}
The function $G:\mathbb{R} \mapsto \mathbb{R}$ in Eqn. \eqref{eq:work_loss} is twice-differentiable and strongly convex on $[-\gamma, \gamma]$, where $\gamma$ is the upper bound in Assumption \ref{assumption: bound_of_Theta}, i.e., there exist two constants $l_\gamma$, $u_\gamma > 0$ that satisfy
\begin{equation}
\nonumber
l_\gamma \leq G''(x) \leq u_\gamma, ~ \forall x\in [-\gamma, \gamma].
\end{equation} 
\end{assumption}

\begin{remark}
Assumption \ref{assumption: bounds_of_ddG} implies that the Bregman divergence of $\Phi_{t}$ at \eqref{eq:work_loss} satisfies that for any $x, y \in [-\gamma, \gamma]$, we have
\begin{equation}
\nonumber
l_\gamma (x-y)^2/2
\le B_{\Phi_{t}}(x,y)
\le u_\gamma (x-y)^2/2.
\end{equation} For example, if the noise in \eqref{eq:regression_model} follows a Gaussian distribution $\mathcal{N}(0, \sigma^2)$, then two convexity constants are equal to the standard deviation, i.e., $l_\gamma=u_\gamma=\sigma$. Table \ref{table:log_partition} gives more concrete examples for $G(x)$ that satisfy Assumption \ref{assumption: bounds_of_ddG}.
\end{remark}

The noise sequence $\{\epsilon_{t}\}_{t=1}^{T}$ in \eqref{eq:regression_model} accounts for the randomness that arises from the sampling process. Different from \citet{lafond2015online} which assumes that the noises $\{\epsilon_{t}\}_{t=1}^{T}$ are drawn independently and identically from a fixed distribution, we consider a more realistic dependent noise sequence drawn from a martingale difference sequence that is adapted to the history, i.e., with respect to a $\sigma-$field 
\begin{equation}
\label{eq: filtration}
    \mathcal{H}_{t-1} = \sigma\left( \pi_1, y_1, \pi_2, y_2, \cdots, \pi_{t-1}, y_{t-1}, \pi_t \right)
\end{equation}
generated by all previous samples $\{\pi_{s}\}_{s=1}^{t}$ and all previous feedback $\{y_{s}\}_{s=1}^{t-1}$ before the feedback $y_{t}$ is observed, the noise process $\{\epsilon_{t}\}_{t=1}^{T}$ satisfies $\mathbb{E}[\epsilon_{t}|\mathcal{H}_{t-1}]=0$ for all time point $t \in [T]$.

We consider two class of  non-parametric noise sequences $\{\epsilon_{t}\}_{t=1}^{T}$ in \eqref{eq:regression_model}: sub-Gaussian martingale difference and sub-Exponential martingale difference with respect to the history $\sigma-$ field $\mathcal{H}_{t-1}$ at \eqref{eq: filtration}. Formally,

\begin{definition}\label{def:sub-Gau}
\textbf{\upshape (Sub-Gaussian Martingale Difference Noise Process)} A noise process $\{\epsilon_{t}\}_{t=1}^{T}$ is called a sub-gaussian martingale difference noise process with variance proxy $\sigma^{2}$ if for all $t \in \{1,\cdots, T\}$, 
\begin{equation}
\mathbb{E}[\exp(s\epsilon_{t})|\mathcal{H}_{t-1}] \leq \exp(s^2\sigma^2/2), \forall s \in \mathbb{R}.
\end{equation} 
\end{definition}

\begin{definition}\label{def:sub-Exp}
\textbf{\upshape (Sub-Exponential Martingale Difference Noise Process)} 
A noise process $\{\epsilon_{t}\}_{t=1}^{T}$ is called a sub-exponential martingale difference noise process with parameter $\lambda$ if for all $t \in \{1,\cdots, T\}$, 
\begin{equation}
\mathbb{E}[\exp(s\epsilon_{t})|\mathcal{H}_{t-1}] \leq \exp(s^2\lambda^2/2), \forall |s| \le 1/\lambda.
\end{equation}
\end{definition}






\section{Formal results}
\label{sec:theory_results}

This section presents the formal results towards building the always-valid prediction risk bound process (Definition \ref{def:AVRB}).
Section \ref{subsec:penal_design} presents the key connection between the regularization sequence and prediction error sequence. 
Section \ref{subsec:gradient_control} presents a key application of non-asymptotic martingale concentration to deliver always-valid gradient norm control. Section \ref{subsec:av_pred_proc} presents the resulting always-valid prediction risk bound process.

\subsection{Penalization design principle}
\label{subsec:penal_design}

This section explains why the regularization sequence $\{\lambda_{t}\}_{t=1}^{T}$ is designed as in Lemma \ref{lem: choice_of_lambda_subgaussian} for sub-Gaussian and in Lemma \ref{lem: choice_of_lambda_subexponential} for sub-Exponential martingale difference noise processes and how the upper bounds in Theorem \ref{thm: oracle_inequality} are derived. 

\textbf{Choice of Regularization}. Intuitively, the optimal choice of regularization sequence relies on one's art of bias-and-variance trade-off. Bias arises as a shrinkage effect from nuclear norm regularizer in \eqref{eq:work_program} and grows as the regularization level $\lambda_{t}$ increases. Besides, the regularization sequence $\{\lambda_{t}\}_{t=1}^{T}$ should offset the fluctuations in the score function process $\{\nabla \Phi_t(\mathbf{\Theta})\}_{t=1}^{T}$, formally defined as 
\begin{equation}
\label{eq:work_score}
    \begin{aligned}
       \nabla \Phi_t(\mathbf{\Theta}) &= \frac{1}{t} \sum_{i=1}^t  \bigg(  G'(\TrInnProd{\mathbf{E}_{\pi_i}}{\mathbf{\Theta}}) - \Obs_i \bigg) \mathbf{E}_{\pi_i} 
    \end{aligned}
\end{equation}
for the loss function process \eqref{eq:work_loss}. Consequently, an optimal choice of the regularization sequence $\{\lambda_{t}\}_{t=1}^{T}$ is the smallest \textit{envelop} that is large enough and \textit{always} controls the fluctuations of score function process $\{\nabla \Phi_t(\mathbf{\Theta})\}_{t=1}^{T}$ along the whole online learning process. 

Precisely, our goal is to design a regularization sequence $\{\lambda_{t}(\alpha)\}_{t=1}^{T}$ such that, for a given confidence budget $\alpha \in (0,1)$, it holds that 
\begin{equation}\label{eq:AV_good_event}
\mathbb{P}_{\mathbf{\Theta}^*}\bigg(
\mathfrak{G}\left(\left\{\lambda_{t}(\alpha) \right\}_{t=1}^{T}\right)
\bigg) \ge 1-\alpha, 
\end{equation}
The event $
\mathfrak{G}(\left\{\lambda_{t}\right\}_{t=1}^{T})= \{\forall t \in [T]: G_{t} \text{ holds}\}$ describes a sequence of good events $\{G_{t}\}_{t=1}^{T}$. The good event $G_{t}$ is defined as
\begin{equation}\label{eq:good_event}
G_t :=\left\{  {\lambda_{t}} \geq 2 \vertiii{\frac{1}{t} \sum_{i=1}^{t}
\bigg(G'(\TrInnProd{\mathbf{E}_{\pi_i}}{\mathbf{\Theta}}) - \Obs_i \bigg) \mathbf{E}_{\pi_i}}_{op} \right\},
\end{equation} 

Consequently, given that the agent learns the target matrix parameter $\TruePara$ by solving the program \eqref{eq:work_program} with the specified regularization scheme (\eqref{eq:subGau_reg} for sub-Gaussian noise and  \eqref{eq:subExp_reg} for sub-Exponential noise), the resulting estimator process $\{\widehat{\mathbf{\Theta}}_{1}, \widehat{\mathbf{\Theta}}_{2}, \cdots, \widehat{\mathbf{\Theta}}_{T}\}$ enjoys an \textit{always-validity}, that is, the online statistical learning procedure is theoretically valid at \textit{every} time point with a time-uniform estimation error bound. Such always-validity is an ideal property for online regularized statistical learning
and serves as a warranty on the robustness of online learning algorithm design.

\textbf{Restricted Strong Convex condition.} 
To derive an envelop for the estimate error random walk $\{\vertiii{\mathbf{\widehat{\Theta}}_t - \mathbf{\Theta^*}}_{F}^{2} \}_{t=1}^{T}$ of the minimizer sequence $\{\mathbf{\widehat{\Theta}}_t\}_{t=1}^{T}$ of low rank regularized problem \eqref{eq:work_program}, we introduce a process level condition that extends the  restricted strong convex condition in high-dimensional statistical estimation literature \cite{wainwright2019high}:
\begin{definition}
\label{def: restricted_strong_convexity}

For any $\mathbf{\Theta}, \mathbf{\Theta}^*$ in $\mathbb{R}^{d_1 \times d_2}$, a function $\Phi_t(\mathbf{\Theta})$ is said to be $\kappa_{t}$ \textbf{restricted strong convex} with respect to the Frobenius norm $\vertiii{\cdot}$ and with tolerance $\tau^2(\mathbf{\Theta}^*)$ if its Bregman divergence satisfies the that, $\forall \mathbf{\Theta} \in \mathbb{R}^{p_{1} \times p_{2}}$,
\begin{equation}\label{eq:condi_rsc}
    B_{\Phi_t}(\mathbf{\Theta}, \mathbf{\Theta^*})  \geq \kappa_t \vertiii{\mathbf{\Theta}- \mathbf{\Theta^*}}_{F}^{2}-\tau_{t}^{2}(\mathbf{\Theta^*}).
\end{equation}
\end{definition}

Given the restricted strong convex condition \eqref{eq:condi_rsc}, one has a estimation error bound as follows:
\begin{lemma}
\label{lem: oracle_inequality}
 \textbf{\upshape(Estimation error bound under RSC, Proposition 10.6 in \citet{wainwright2019high})}. Suppose that at time $t$, the observations $\mathbf{E}_{\pi_1}, \ldots, \mathbf{E}_{\pi_t}$ satisfies the non-scaled RSC condition in Definition \eqref{def: restricted_strong_convexity}, such that the the $\kappa_{t}-$restricted strong convex condition \eqref{eq:condi_rsc}.
Then under the good event \eqref{eq:good_event}
the optimal solution $\widehat{\mathbf{\Theta}}_t$ to the regularized loss function $\Phi_t(\mathbf{\Theta}) + \lambda_t \vertiii{\mathbf{\Theta}}_{nuc}$ satisfies the bound below:
\begin{equation}\label{eq:rsc_oracle}
\vertiii{\mathbf{\widehat{\Theta}}_t - \mathbf{\Theta^*}}_{F}^{2} \leq 9 \frac{\lambda_{t}^{2}}{\kappa_t^{2}}r + \frac{2}{\kappa_t}\tau_t^2(\mathbf{\Theta^*})
\end{equation}
\end{lemma}
 
In our case, the tolerance is zero ($\tau_t^2(\mathbf{\Theta^*}) = 0)$ and $\kappa_{t} = \bar{p}_{t}$ at \eqref{eq:min_samp_frequency}. These condition reduce \eqref{eq:rsc_oracle} into 
\begin{equation}
\vertiii{\mathbf{\widehat{\Theta}}_t - \mathbf{\Theta^*}}_{F}^{2} \leq 9 \frac{\lambda_{t}^{2}}{\bar{p}_{t}^{2}}r
\end{equation}
Consequently, the convergence rate depends on two important component: the minimal empirical sampling frequency $\bar{p}_{t}$ at \eqref{eq:min_samp_frequency} and the regularization level $\lambda_{t}$ at \eqref{eq:Ref_Process}.

\subsection{Gradient norm control}
\label{subsec:gradient_control}

\textbf{Always-Valid Control of Gradient Operator Norm.}
Here we show how to design an online regularization sequence to tame sequence of good event \eqref{eq:good_event}. We establish two choices of the regularization sequence $\{\lambda_t\}_{t=1}^{T}$, for both sub-Gaussian and sub-Exponential martingale difference, respectively.

\begin{lemma}
\label{lem: choice_of_lambda_subgaussian} 
\textbf{\upshape (sub-Gaussian case)}
Suppose that $\{\epsilon_t\}_{t=1}^\infty$ is conditional $\sigma^2$-sub-Gaussian. For any $\alpha \in (0,1)$, set $\lambda_t$ as
\begin{equation}
\label{eq:subGau_reg}
    \lambda_t = 8\sigma\sqrt{\frac{\log ((d_1 + d_2)/\alpha) S_{t}}{t}}
\end{equation}
then the event \eqref{eq:AV_good_event} (that is, $\lambda_t \geq  2\vertiii{ \nabla \Phi_{t}(\mathbf{\Theta}^*)}_{op}, \forall t \in [T]$) holds with probability at least $1-\alpha$. 
\end{lemma}

Lemma \ref{lem: choice_of_lambda_subgaussian} indicates that, if the noise in the feedback generating mechanism 
\eqref{eq:regression_model} is of conditional $\sigma-$subGaussian distribution (Definition \ref{def:sub-Gau}), then the event \eqref{eq:AV_good_event} holds with probability at least $1-\alpha$ by choosing the regularization sequence $\{\lambda_{t}\}_{t=1}^{T}$ with \eqref{eq:subGau_reg}. Plug such choice of $\{\lambda_{t}\}_{t=1}^{T}$ into \eqref{eq:rsc_oracle} in lemma \ref{lem: oracle_inequality}, we have the always valid prediction risk bound process \eqref{eq:EstErr_subGau} presented at the main Theorem \ref{thm: oracle_inequality}.

\begin{lemma}
\label{lem: choice_of_lambda_subexponential} 
\textbf{\upshape (sub-Exponential case)} Suppose that $\{\epsilon_t\}_{t=1}^\infty$ is conditional $\lambda$-sub-Exponential.
For any $\alpha \in (0,1)$, set $\lambda_t$ as
\begin{equation}
\label{eq:subExp_reg}
    \lambda_t =4 8\lambda \sqrt{\frac{S_t \log ((d_1 + d_2)/\alpha)}{t}} + 48\lambda \frac{\log ((d_1 + d_2)/\alpha)}{t} 
\end{equation}
then the event \eqref{eq:AV_good_event} (that is, $\lambda_t \geq  2\vertiii{ \nabla \Phi_{t}(\mathbf{\Theta}^*)}_{op}, \forall t \in [T]$) holds with probability at least $1-\alpha$. 
\end{lemma}

Lemma \ref{lem: choice_of_lambda_subexponential} indicates that, if the noise in the feedback generating mechanism 
\eqref{eq:regression_model} is of conditional $\lambda-$subExponential distribution (Definition \ref{def:sub-Exp}), then the event \eqref{eq:AV_good_event} holds with probability at least $1-\alpha$ by choosing the regularization sequence $\{\lambda_{t}\}_{t=1}^{T}$ with \eqref{eq:subExp_reg}. Plug such choice of $\{\lambda_{t}\}_{t=1}^{T}$ into \eqref{eq:rsc_oracle} in lemma \ref{lem: oracle_inequality}, we have the always valid prediction risk bound process \eqref{eq:EstErr_subEXP} presented at the main Theorem \ref{thm: oracle_inequality}.

\begin{remark}
Proof of the above two lemmas are provided in Appendix \ref{app:main}. We remark that Lemma \ref{lem: choice_of_lambda_subgaussian} and \ref{lem: choice_of_lambda_subexponential} rely on  martingale difference concentration inequalities for products of general matrices and sub-Gaussian and sub-Exponential noises, which advances prior art in \cite{Tropp2011User}. The proofs of these two tools (Section \ref{app:subGaussian}, \ref{app:subExponential}) are novel.
\end{remark}

\subsection{Always-valid prediction error process}
\label{subsec:av_pred_proc}

In this section, we show the main result on constructing the always-valid risk bound process that satisfy \eqref{eq:EP_AlwaysValid_RiskProcess}:
\begin{theorem}
\label{thm: oracle_inequality} (\textbf{Always-Valid Risk Bound Process})
Suppose that Assumption \ref{assumption: bound_of_Theta}, \ref{assumption: bounds_of_ddG} are satisfied. Then we have the following always-valid prediction risk bound process for sub-Gaussian 
(Definition \ref{def:sub-Gau}) and sub-Exponential (Definition \ref{def:sub-Exp}) martingale difference noise process
, respectively.

\textbf{(Sub-Gaussian case)}. Suppose the noise process $\{\epsilon_t\}_{t=1}^{T}$ is $\sigma^2$-sub-Gaussian
martingale difference
(Definition \ref{def:sub-Gau}), then, under the online regularization  $\{\lambda_t\}_{t=1}^{T}$ as in Lemma \ref{lem: choice_of_lambda_subgaussian}, with probability at least $1-\alpha$, the following inequality holds 
\begin{equation}
\begin{aligned}
\label{eq:EstErr_subGau}
\nonumber
\forall t>0, \vertiii{\mathbf{\widehat{\Theta}}_t - \mathbf{\Theta^*}}_{F}
 \leq \frac{48 \sigma \sqrt{r} }{\bar{p}_t l_{\gamma}} \sqrt{\frac{ S_t  \log ((d_1 + d_2)/\alpha)  }{t}} 
\end{aligned}
\end{equation}

\textbf{(Sub-Exponential case)}. Suppose the noise process $\{\epsilon_t\}_{t=1}^{T}$ is $\lambda$-sub-Exponential
martingale difference
(Definition \ref{def:sub-Exp}), then, under the online regularization 
 $\{\lambda_t\}_{t=1}^{T}$ as in  Lemma \ref{lem: choice_of_lambda_subexponential}, with probability at least $1-\alpha$, the following inequality holds
\begin{equation}
\label{eq:EstErr_subEXP}
\begin{aligned}
\nonumber
\forall t>0, &\vertiii{\mathbf{\widehat{\Theta}}_t - \mathbf{\Theta^*}}_{F}
 \leq \frac{288\lambda \sqrt{r}}{\bar{p}_t l_{\gamma}} 
 \left( \sqrt{\frac{S_t \log ((d_1 + d_2)/\alpha)}{t}}  + \frac{\log ((d_1 + d_2)/\alpha)}{t}  \right)
 \end{aligned}
\end{equation}
\end{theorem}

\begin{remark}
The above oracle inequality is always-valid, which means that the upper bound is valid with high probability for all $t > 0$ simultaneously. Ignore the time-dependent parameters $\bar{p}_t$ and $S_t$ at the moment, then the above upper bound of $\vertiii{\mathbf{\Theta_t} - \mathbf{\Theta}^*}_{F}$ matches the best-known upper bound in the offline case, i.e., $\widetilde{\mathcal{O}}({1}/{\sqrt{t}})$ \citep{wainwright2019high}.

\end{remark}

\begin{remark}
The two summary statistics,  $\bar{p}_t$ at \eqref{eq:min_samp_frequency} and $S_t$ at \eqref{eq: Sigma} depend on the adopted sampling policy $\Pi$ of the matrix elements.
For example, if the sampling policy is close to a uniform sampling after some $T_0 > 0$, then we would have
\begin{equation}
\tag*{(C-1)}
\label{eq: condition_p_t}
\begin{aligned}
\forall t > T_0, \bar{p}_t \geq 1/ (\mu d_1 d_2), \text{ for some } \mu \geq 1
 \end{aligned}
\end{equation}
which means that the sampling probability of each element (there are $d_1 d_2$ elements) is lower bounded by a constant. Also, if we denote $R_t^k = \sum_{l=1}^{d_2} p_t(k,l)$, $C_t^l = \sum_{k=1}^{d_1} p_t(k,l)$ to be the probability of the sampling from row $k$ and column $l$ respectively. Another concequence for a uniform sampling policy is that no row nor column is sampled far more frequently than the others, i.e.,
\begin{equation}
\tag*{(C-2)}
\begin{aligned}
\label{eq: condition_S_t}
\forall t > T_0, \max(R_t^k, C_t^l) \leq \frac{\nu}{d_1 \wedge d_2} \text{ for some } \nu \geq 1
 \end{aligned}
\end{equation}
%
These conditions are often assumed in the offline matrix completion problems \citep{wainwright2019high, lafond2015low}, and we only use them to show in the following corollary that, in terms of the dimensions $d_1, d_2$ and the time $t$, the asymptotic rate of our always-valid bound matches the best-known Hoeffding-type bound in the offline setting if given the same assumptions.

\end{remark}

\begin{corollary}
Suppose that the noise process is conditional $\sigma^2$-sub-Gaussian. Further suppose that Conditions \ref{eq: condition_p_t} \ref{eq: condition_S_t} are satisfied. Then we have $\forall t>T_0>0$, with probability at least $1-\alpha$
\begin{equation}
\begin{aligned}
\nonumber
\vertiii{\mathbf{\widehat{\Theta}}_t - \mathbf{\Theta^*}}_{F}
 \leq \frac{48d_1 d_2 \mu \sigma \sqrt{r} }{ l_{\gamma}} \sqrt{\frac{ \nu u_{\gamma}^2 \log ((d_1 + d_2)/\alpha)  }{(d_1 \wedge d_2) t}} 
\end{aligned}
\end{equation}
\end{corollary}

\begin{remark}
The above rate matches the best-known convergence rate of the Hoeffding-type upper bounds on the Frobenius norm in the offline matrix completion problem, i.e., $\widetilde{\mathcal{O}}\left((d_1^2d_2 \vee d_1d_2^2 )^{1/2}/\sqrt{t})\right)$ by \citet{lafond2015low}. Similar results can be obtained for the sub-Exponential case by combining Conditions \ref{eq: condition_p_t} \ref{eq: condition_S_t} with Eqn. \eqref{eq:EstErr_subGau}. 
\end{remark}

\section{Experiments}
\label{sec: experiments}
\begin{figure*}[ht!]
\centering
\hspace*{-1.7em}
\subfigure[\footnotesize Gaussian]{
  \centering
  \includegraphics[width=0.33\linewidth]{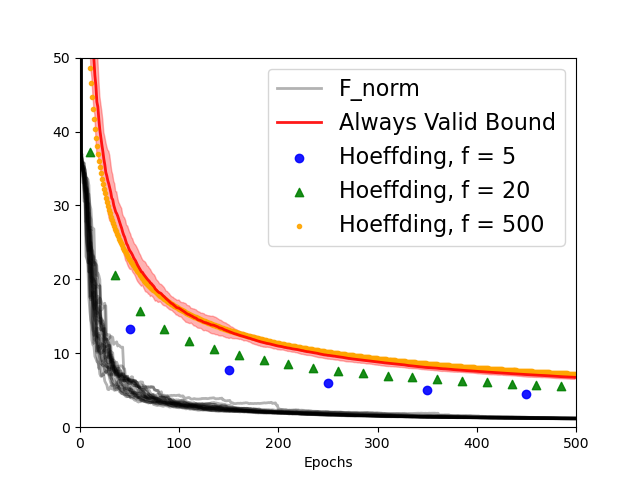}
  \label{fig: Gaussian}
}\hspace*{-1.7em}%
\subfigure[\footnotesize Binomial]{
  \centering
  \includegraphics[width=0.33\linewidth]{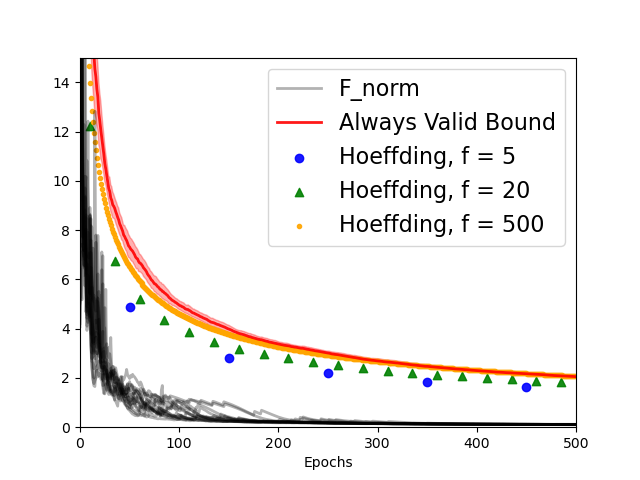}
    \label{fig: Binomial}
}\hspace*{-1.7em}%
\subfigure[\footnotesize Poisson ]{
  \centering
  \includegraphics[width=0.33\linewidth]{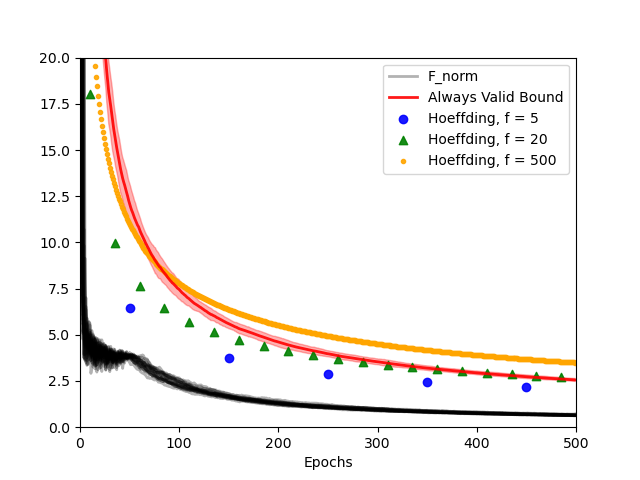}
  \label{fig: Poisson}
}
\caption{ Experimental results on the synthetic $5\times 5$ matrices. F-norm denotes the Frobenius norm of the difference $\mathbf{\Theta}_t - \mathbf{\Theta}^*$ between the estimator and the parameter. (f : number of check points in Hoeffding bounds)}
\label{fig: experimental_results_5}
\end{figure*}

\begin{figure*}[ht!]
\centering
\hspace*{-1.7em}
\subfigure[\footnotesize Gaussian]{
  \centering
  \includegraphics[width=0.33\linewidth]{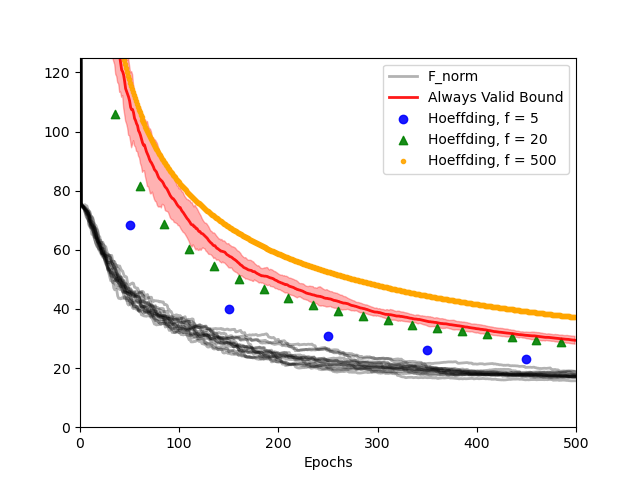}
  \label{fig: Gaussian10}
}\hspace*{-1.7em}%
\subfigure[\footnotesize Binomial]{
  \centering
  \includegraphics[width=0.33\linewidth]{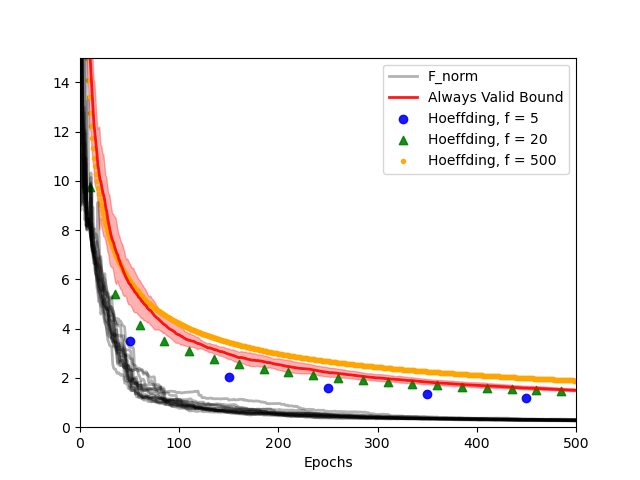}
    \label{fig: Binomial10}
}\hspace*{-1.7em}%
\subfigure[\footnotesize Poisson ]{
  \centering
  \includegraphics[width=0.33\linewidth]{new_figs/Poisson10.png}
  \label{fig: Poisson10}
}
\caption{ Experimental results on the synthetic $10\times 10$ matrices. F-norm denotes the Frobenius norm of the difference $\mathbf{\Theta}_t - \mathbf{\Theta}^*$ between the estimator and the parameter. (f: number of check points in Hoeffding bounds)}
\label{fig: experimental_results_10}
\end{figure*}

In this section, we validate our main theoretical results of the always-valid prediction risk bound process (Theorem \ref{thm: oracle_inequality}) on synthetic data.

\textbf{Synthetic Data Generation Mechanism.} We choose three generalized linear models as examples of our general theory: Gaussian, Binomial and Poisson with their corresponding $G(x)$ as in Table \ref{table:log_partition}. For all the models, we generate the target matrices $\mathbf{\Theta}^*$ by first generating $r$ vectors randomly, and then each row of the target matrices chooses either zero or one of the generated vectors uniform randomly so that every $\mathbf{\Theta}^*$ is low-rank. We consider two settings where the target matrices have dimensions $(d_1, d_2) = (5, 5), (10, 10)$ and rank $r=1, 2$ respectively. For each model, we run the matrix completion procedure for 20 independent runs.
More experimental details can be found in Appendix \ref{app:experiments}.

\textbf{Figure \ref{fig: experimental_results_5}, \ref{fig: experimental_results_10}: Risk monitoring under Gaussian, binomial and Poisson online matrix completion.} The Frobenius norm of the difference $\mathbf{\Theta}_t - \mathbf{\Theta}^*$, the always valid upper bounds, and the Hoeffding upper bounds with pre-specified checkpoints in these two settings are shown in Figure\ref{fig: experimental_results_5}, \ref{fig: experimental_results_10} respectively. To be more specific, for Gaussian and Binomial, we plot their always-valid upper bounds in Eqn. \eqref{eq:EstErr_subGau} as they are sub-Gaussian. For Poisson, we plot its always-valid upper bounds in Eqn. \eqref{eq:subExp_reg} as it is sub-Exponential. For the Hoeffding-type of upper bound, we choose to plot the upper bounds in \citet{lafond2015low} with the same overall confidence as always-valid bounds. 

Figure \ref{fig: experimental_results_5} and \ref{fig: experimental_results_10} indicate that both the Hoeffding upper bounds and the always-valid upper bound correctly measure the change of the prediction error when the number of samples increase. However, the Hoeffding upper bounds become very loose bounds once we add the number of checkpoints. Such a failure is unavoidable since Hoeffding upper bounds only holds for a given \textit{fixed} time step, while always-valid results hold at \textit{every} time point simultaneously. To meet the same confidence budget over a large number of checkpoints, Hoeffding bounds are forced to commit small confidence at each checkpoint while always-valid bounds is free from such issues.


\section{Conclusions and Future Directions}
\label{sec: conclusions}

We have proposed a framework to establish the always-valid concentration inequality for the continuous monitoring online matrix completion problem with non-parametric martingale difference noise process. In terms of online regularization sequence design, we derive novel non-asymptotic matrix martingale concentration to deliver continuous control on the operator norm of the gradient process with respect to online optimization procedure for matrix completion. Based on the online regularization sequence, we establish always valid risk bound process for online matrix completion. The convergence rate of the proposed procedure matches the best-known rate in offline matrix completion.

In the end, we highlight a few particularly relevant questions that are left as future works. The first one is a more general strategy to design online regularization for bandit problems. Note that we only consider \textit{uniform-sampling} based policy in current studies, which takes its advantage on the independence between policy and feedback in our analysis. In bandit or actively learning setting, the policy may actively design which matrix element to observe in next step to gain further sample efficiency and enjoy better convergence rate. 

The second one is more relevant to learning model: can we extend current framework to more general matrix regression setting with more general non-parametric class of martingale difference noise sequence? Beyond the low-rank matrix completion we studied in this paper, general matrix regression includes many important tasks such as multivariate regression, low-rank matrix compressed sensing, phase retrieval, time-series and vector autoregressive processes. It is an exciting direction on how to provide a unified framework for always valid matrix regression to cover all tasks mentioned above, which is important and of demand on online statistical inference from modern online service industry.

\nocite*{}

\bibliography{myref}
\bibliographystyle{plainnat}

\onecolumn
\appendix

\section{Main Proofs}
\label{app:main}
\subsection{Notations}

We start the proof with clarifying all the notations used throughout this section. Note that the maximum log likelihood loss function to be minimized at each iteration is

\begin{equation}
    \begin{aligned}
        \Phi_t(\mathbf{\Theta}) = - \frac{1}{t} \sum_{i=1}^t \left [\log (h(\Obs_i)) + \TrInnProd{\mathbf{E}_{\pi_i}}{\mathbf{\Theta}} \Obs_i - G(\TrInnProd{\mathbf{E}_{\pi_i}}{\mathbf{\Theta}})\right]
    \end{aligned}
\end{equation}

and the algorithm optimizes $\Phi_t(\Theta) + \lambda_t \vertiii{\Theta^*}_{nuc}$. Then the gradient of $\Phi_t(\Theta)$, $\nabla \Phi_t(\Theta)$ is

\begin{equation}
    \begin{aligned}
       \nabla \Phi_t(\mathbf{\Theta}) &= \frac{1}{t} \sum_{i=1}^t \left [  \left (  G'(\TrInnProd{\mathbf{E}_{\pi_i}}{\mathbf{\Theta}}) - \Obs_i \right) \mathbf{E}_{\pi_i} \right] = \frac{1}{t} \sum_{i=1}^t  \epsilon_i \mathbf{E}_{\pi_i}
    \end{aligned}
\end{equation}

where we have defined $\epsilon_i = G'(\TrInnProd{\mathbf{E}_{\pi_i}}{\mathbf{\Theta}}) - \Obs_i$ for each $i$. The Bregman divergence of $\Phi_t$ is

\begin{equation}
    \begin{aligned}
     B_{\Phi_t}(\mathbf{\Theta},  \mathbf{\Theta^*}) 
     &=  \Phi_t(\mathbf{\Theta}) - \Phi_t(\mathbf{\Theta^*}) - \TrInnProd{\nabla \Phi_t(\mathbf{\Theta})}{\mathbf{\Theta} - \mathbf{\Theta}^*}\\
     &=  \frac{1}{t} \sum_{i=1}^t \left [      \TrInnProd{\mathbf{E}_{\pi_i}}{\mathbf{\Theta^*} - \mathbf{\Theta}} \Obs_i - G(\TrInnProd{\mathbf{E}_{\pi_i}}{\mathbf{\Theta}^*}) + G(\TrInnProd{\mathbf{E}_{\pi_i}}{\mathbf{\Theta}}) \right.\\
     &\quad \left. + \left (  G'(\TrInnProd{\mathbf{E}_{\pi_i}}{\mathbf{\Theta}}) - \Obs_i \right)  \TrInnProd{\mathbf{E}_{\pi_i} }{ \mathbf{\Theta}^* - \mathbf{\Theta}}\right] \\
    &=  \frac{1}{t} \sum_{i=1}^t \left [    \left (  G'(\TrInnProd{\mathbf{E}_{\pi_i}}{\mathbf{\Theta}}) \right)  \TrInnProd{\mathbf{E}_{\pi_i} }{ \mathbf{\Theta}^* - \mathbf{\Theta}}   - G(\TrInnProd{\mathbf{E}_{\pi_i}}{\mathbf{\Theta}^*}) + G(\TrInnProd{\mathbf{E}_{\pi_i}}{\mathbf{\Theta}})  \right]\\
    &= \frac{1}{t} \sum_{i=1}^t B_{G} ((\TrInnProd{\mathbf{E}_{\pi_i}}{\mathbf{\Theta}}), (\TrInnProd{\mathbf{E}_{\pi_i}}{\mathbf{\Theta^*}}) )
        \end{aligned}
\end{equation}

We define the following term $S_t$ at each iteration in our proof, which is the same as in the main text.

\begin{equation}
{S}_t = \max \left \{\vertiii{\frac{1}{t}\sum_{i=1}^t \mathbf{E}_{\pi_i}  \mathbf{E}_{\pi_i}^T}_{op}, \vertiii{\frac{1}{t} \sum_{i=1}^t \mathbf{E}_{\pi_i}^T \mathbf{E}_{\pi_i}}_{op}\right \}
\end{equation}

\subsection{Proof of Main Theorem \ref{thm: oracle_inequality}}
\textbf{Proof.} 
Note that Assumption \ref{assumption: bounds_of_ddG} implies that $l_\gamma (x-y)^2 \leq 2B_G(x, y) \leq u_\gamma(x-y)^2 $ and from the notation of $\bar{p}_t = \min_{k\in[p_1], l \in [p_2]} \frac{1}{t} \sum_{i=1}^t \mathbbm{1} \{\pi_i = (k,l)\}$, we have the following inequality on the Bregman divergence of $\Phi_t$.

\begin{equation}
    \begin{aligned}
     B_{\Phi_t}(\mathbf{\Theta},  \mathbf{\Theta^*}) 
     &\geq \frac{l_{\gamma}}{2t} \sum_{i=1}^t  \TrInnProd{\mathbf{E}_{\pi_i}}{\mathbf{\Theta} - \mathbf{\Theta}^*}^2 \geq \frac{\bar{p}_t l_\gamma}{2} \vertiii{\mathbf{\Theta} - \mathbf{\Theta}^* }_F^2
    \end{aligned}
\end{equation}

which implies that $\Phi_t(\mathbf{\Theta})$ is $(\bar{p}_t l_{\gamma})$-restricted strong convex with zero tolerance. Now by Lemma \ref{lem: oracle_inequality}, we can obtain the oracle inequality as follows.

\begin{equation}
\begin{aligned}
\vertiii{\mathbf{\widehat{\Theta}}_t - \mathbf{\Theta^*}}_{F}^{2} 
&\leq 9 \frac{\lambda_{t}^{2}}{\kappa_t^{2}}r + \frac{2}{\kappa_t}\tau_t^2(\mathbf{\Theta^*}) 
\end{aligned}
\end{equation}

 By Lemma \ref{lem: choice_of_lambda_subgaussian}, when the noise term $\epsilon_i = G'(\TrInnProd{\mathbf{E}_{\pi_i}}{\mathbf{\Theta}}) - \Obs_i$ is conditional $\sigma^2$-sub-Gaussian, we have that with probability at lesast $1-\alpha$, for all $t \in [T]$,

\begin{equation}
\begin{aligned}
\vertiii{\mathbf{\widehat{\Theta}}_t - \mathbf{\Theta^*}}_{F}
&\leq \frac{6 \sqrt{r} \lambda_t}{\bar{p}_t l_{\gamma}} \leq \frac{48 \sigma \sqrt{r} }{\bar{p}_t l_{\gamma}} \sqrt{\frac{ S_t  \log ((d_1 + d_2)/\alpha)  }{t}} \\
\end{aligned}
\end{equation}

 By Lemma \ref{lem: choice_of_lambda_subexponential}, when the noise term $\epsilon_i = G'(\TrInnProd{\mathbf{E}_{\pi_i}}{\mathbf{\Theta}}) - \Obs_i$ is conditional $\lambda$-sub-Exponential, we have that with probability at least $1-\alpha$, for all $t\in [T]$,

\begin{equation}
\tag*{$\square$}
\vertiii{\mathbf{\widehat{\Theta}}_t - \mathbf{\Theta^*}}_{F}
\leq \frac{6 \sqrt{r} \lambda_t}{\bar{p}_t l_{\gamma}} \leq \frac{288\lambda \sqrt{r}}{\bar{p}_t l_{\gamma}} \left( \sqrt{\frac{S_t \log ((d_1 + d_2)/\alpha)}{t}} + \frac{\log ((d_1 + d_2)/\alpha)}{t}  \right)
\end{equation}

\subsection{Choice of the Regularization Parameter}

\subsubsection{Proof of Lemma \ref{lem: choice_of_lambda_subgaussian}} \label{sec:choice_lambda_subgaussian}
\textbf{Proof.} The proof of this lemma relies heavily on the always-valid concentration inequality for sum of products of matrices and sub-Gaussian random variables we develop in Appendix \ref{app:subGaussian}. We first show the following simple but important fact. For any two sequence of events $A_1, A_2, \cdots A_T$ and $B_1, B_2, \cdots B_T$. If $\mathbb{P}(B_i) = 1, \forall i \in [T]$, then we have $\mathbb{P} \left( \left( \cap_{i=1}^T A_i \right) \cap \left (\cap_{i=1}^T B_i \right)  \right) = \mathbb{P} \left( \left( \cap_{i=1}^T A_i \right) \right)$. The fact can be proved by showing it for two events $A$, $B$ with $\mathbb{P}(B) = 1$.

\begin{equation}
\mathbb{P} \left(A \cap B\right) = \mathbb{P}(A) - \mathbb{P} \left(A \cap B^c\right) \geq \mathbb{P}(A) - \mathbb{P} \left(B^c\right) = \mathbb{P}(A)
\end{equation} 

The proof now follows from taking $A_t = \{\lambda_t \geq  2\vertiii{\frac{1}{t}\sum_{i=1}^t \mathbf{E}_{\pi_i} \epsilon_i }_{op} \}$ and $B_t = \{S_t \leq S_t\}$. Note that by Lemma \ref{lem: always_valid_inequality_rectangular_subgaussian}, we have the following concentration inequality

\begin{equation}
\begin{aligned}
    \mathbb{P}\left (\forall t \in [T]: \lambda_t \geq  2\vertiii{\frac{1}{t}\sum_{i=1}^t \mathbf{E}_{\pi_i} \epsilon_i }_{op} \right ) \geq 1 - (d_1+ d_2) \cdot  \mathrm{exp} \left(- \frac{\sqrt{2} t^2 \lambda_t^2 }{64\sigma^2 t {S}_t }\right)
\end{aligned}
\end{equation}

Let the right-hand-side be $1- \alpha$ and solve for $\lambda_t$, we get the choice of $\lambda_t$ in our algorithm.

\begin{equation}
\tag*{$\square$}
\begin{aligned}
    \lambda_t = \sqrt{\frac{\log ((d_1 + d_2)/\alpha)32\sqrt{2}\sigma^2 S_t}{t}} \leq  8\sigma\sqrt{\frac{\log ((d_1 + d_2)/\alpha) S_t}{t}}
\end{aligned} 
\end{equation}

\subsubsection{Proof of Lemma \ref{lem: choice_of_lambda_subexponential}} \label{sec:choice_lambda_subexp}
\textbf{Proof}. The proof of this lemma relies heavily on the always-valid concentration inequality for sum of products of matrices and sub-Exponential random variables we develop in Appendix \ref{app:subExponential}.
The argument is similar to the proof of Lemma \ref{lem: choice_of_lambda_subgaussian}. Note that each matrix $\mathbf{E}_{\pi_i}$ has only one entry that is one and the other entries are 0, therefore the maximum operator norm of $\mathbf{E}_{\pi_i}$ is 1. By Lemma \ref{lem: always_valid_inequality_rectangular_subexponential}, we know that

\begin{equation}
\mathbb{P}\left (\forall t \geq 0: \lambda_t \geq 2 \vertiii{ \frac{1}{t}\sum_{i=1}^t \mathbf{E}_{\pi_i} \epsilon_i} \right ) \geq \left \{
\begin{aligned}
 & 1 - (d_1 + d_2) \cdot  \mathrm{exp} \left(- \frac{\lambda_t^2 t^2}{2304\lambda^2 t S_t }\right) &\text{ if $\lambda_t \leq 48 \lambda S_t $} \\
 & 1 - (d_1 + d_2) \cdot  \mathrm{exp} \left(- \frac{\lambda_t t}{48\lambda x_{\max}}\right) & \text{ if $\lambda_t \geq 48\lambda  S_t $}
\end{aligned}
\right.
\end{equation}

Therefore let the right hand side be $1-\alpha$ in the two cases and solve for $\lambda_t$, we get that

\begin{equation}
\lambda_t = \left\{
\begin{aligned}
 & 48\lambda \sqrt{\frac{S_t \log ((d_1 + d_2)/\alpha)}{t}} &\text{ if $S_t \geq \frac{\log ((d_1 + d_2)/\alpha)}{t}  $} \\
 & 48\lambda \frac{\log ((d_1 + d_2)/\alpha)}{t} 
 & \text{  if $S_t \leq \frac{\log ((d_1 + d_2)/\alpha)}{t}  $}
\end{aligned}
\right.
\end{equation}

Hence if we take $\lambda_t$ to be the sum (or the maximum) of the above two choices, we can obtain that

\begin{equation}
\tag*{$\square$}
\mathbb{P}\left (\forall t \geq 0: \lambda_t \geq 2 \vertiii{ \frac{1}{t}\sum_{i=1}^t \mathbf{E}_{\pi_i} \epsilon_i} \right ) \geq 1-\alpha
\end{equation}

\subsection{Known Sampling Policy}

\begin{lemma}
\textbf{\upshape(Blackwell Inequality)}
\label{lem: blackwell}
Let $\left(S_t\right)_{t=0}^{\infty}$ be a real-valued martingale with respect to a filtration $\left(\mathcal{F}_t\right)_{t=0}^{\infty}$, with $S_0=0$. If $\left|\Delta S_t\right| \leq 1$ for all $t$, then for any $a, b>0$, we have
\begin{equation}
    \mathbb{P}\left(\exists t \in \mathbb{N}: S_t \geq a+b t\right) \leq e^{-2 a b}
\end{equation}
\end{lemma}

\textbf{Proof.} 
Let $(k,l) = \text{argmin}_{i \in [d_1], j \in [d_2]} p_{i, j}$, substitute $S_t = \sum_{t=1}^T \left( \mathbb{I}(\pi_t =(k,l)) - p_{k,l} \right)$, $a = bt$, and $b = {p_{k,l}}/{4}$, we get the following inequality:

\begin{equation}
    \mathbb{P}\left(\exists t \in \mathbb{N}: S_t \geq - \frac{p_{k,l}t}{2} \right) \leq e^{- \frac{p_{k,l}^2}{8}t}
\end{equation}

Take $\alpha = e^{-p_{k,l}^2/8 t}$, we get $t = \frac{8\log (1/\alpha)}{p^2_{k,l}}$

\begin{equation}
   p_t - \frac{1}{d_1 d_2} \geq - \frac{p}{2}
\end{equation}


\section{Always Valid Inequality for Sum of Matrices and Sub-Gaussians}
\label{app:subGaussian}

In this section, we present the always-valid inequality for the sum of products of matrices and sub-Gaussian random variables, which may be of independent interest to the readers. We prove such inequalities for both self-adjoint matrices and general rectangular matrices. As will be shown in our proof, the latter inequality is based on the former one. 

\subsection{Proof of Lemma \ref{lem: always_valid_inequality_rectangular_subgaussian}}

\begin{lemma}
\label{lem: always_valid_inequality_rectangular_subgaussian}  \textbf{\upshape (Always Valid Inequality for Rectangular Matrices, Sub-Gaussian Case)}
For any sequence of rectangular matrices $\{\mathbf{X}_i\}_{i=1}^{\infty}$ with $\mathbf{X}_i \in \mathbb{R}^{d_1 \times d_2}$ and any sequence of conditional $\sigma^2$-sub-Gaussian noises $\{\epsilon_i\}_{i=1}^{\infty}$, define the following quantity 

\begin{equation}
S_k = \max \left \{\vertiii{\sum_{i=1}^k \mathbf{X}_i  \mathbf{X}_i^T}_{op}, \vertiii{\sum_{i=1}^k \mathbf{X}_i^T \mathbf{X}_i}_{op}\right \}
\end{equation}

Then we have that for any $t > 0$ and $w>0$,
\begin{equation}
\begin{aligned}
\mathbb{P}\left (\exists k \geq 0: \vertiii{\sum_{i=1}^k \mathbf{X}_i \epsilon_i }_{op} \geq t \text { and } S_k \leq w\right ) \leq (d_1+ d_2) \cdot  \mathrm{exp} \left(- \frac{\sqrt{2}t^2}{16\sigma^2 w }\right)
\end{aligned}
\end{equation}

\end{lemma}

\textbf{Proof.} The Lemma essentially proves that Lemma \ref{lem: always_valid_inequality_self_adjoint_subgaussian} also holds for arbitrary rectangular matrices, with a slightly different scaling constant. Given a tail inequality for self-adjoint matrices, we can use the trick of \textit{dilation} to expand any rectangular matrix into a self-adjoint matrix while preserving the operator norm. Define a sequence of self-adjoint matrices $\{\mathbf{Z}_i\}_{i=1}^{\infty}$ with dimension $d = d_1 + d_2$ by
\begin{equation}
\mathbf{Z}_i = 
\begin{bmatrix}
& \mathbf{0} & \mathbf{X}_i \\
& \mathbf{X}_i^T  & \mathbf{0}
\end{bmatrix}
\end{equation}

Then by Lemma \ref{lem: always_valid_inequality_self_adjoint_subgaussian}, we have that for any $t > 0$ and $w>0$,

\begin{equation}
\begin{aligned}
\mathbb{P}\left (\exists k \geq 0: \lambda_{\max }\left(\sum_{i=1}^k \mathbf{Z}_i \epsilon_i \right) \geq t \text { and } \lambda_{\max }\left(\sum_{i=1}^k \mathbf{Z}_i^2 \right) \leq w\right ) \leq (d_1+ d_2) \cdot  \mathrm{exp} \left(- \frac{\sqrt{2}t^2}{16\sigma^2 w }\right)
\end{aligned}
\end{equation}

By Lemma \ref{lem: property_of_dilation}, we know that \footnote{$\sum_{i=1}^k  \mathbf{Z}_i^2 $  is a diagonal matrix, but by an argument similar to Lemma\ref{lem: property_of_dilation}, we can prove the desired property.}

\begin{equation}
\vertiii{\sum_{i=1}^k \mathbf{Z}_i \epsilon_i}_{op} = \vertiii{\sum_{i=1}^k \mathbf{X}_i \epsilon_i}_{op} \text{ and } \vertiii{\sum_{i=1}^k \mathbf{Z}_i^2 }_{op} = \max \left \{\vertiii{\sum_{i=1}^k \mathbf{X}_i^T \mathbf{X}_i  }_{op}, \vertiii{\sum_{i=1}^k \mathbf{X}_i \mathbf{X}_i^T  }_{op} \right \} 
\end{equation}

Therefore we have the following tail inequality 

\begin{equation}
\tag*{$\square$}
\begin{aligned}
\mathbb{P}\left (\exists k \geq 0: \vertiii{\sum_{i=1}^k \mathbf{X}_i \epsilon_i }_{op} \geq t \text { and } S_k \leq w\right ) \leq (d_1+ d_2) \cdot  \mathrm{exp} \left(- \frac{\sqrt{2}t^2}{16\sigma^2 w }\right)
\end{aligned}
\end{equation}

\begin{lemma}
\label{lem: always_valid_inequality_self_adjoint_subgaussian}
\textbf{\upshape (Always Valid Inequality for Self-adjoint Matrices, Sub-Gaussian Case)}
For any sequence of self-adjoint matrices $\{\mathbf{X}_i\}_{i=1}^{\infty}$ with $\mathbf{X}_i \in \mathbb{R}^{d\times d}$ and  any sequence of conditional $\sigma^2$-sub-Gaussian noises $\{\epsilon_i\}_{i=1}^{\infty}$, we have that for any $t > 0$ and $w>0$,
\begin{equation}
\begin{aligned}
\mathbb{P}\left (\exists k \geq 0: \lambda_{\max }\left(\sum_{i=1}^k \mathbf{X}_i \epsilon_i \right) \geq t \text { and } \lambda_{\max }\left(\sum_{i=1}^k \mathbf{X}_i^2 \right) \leq w\right ) \leq d \cdot  \mathrm{exp} \left(- \frac{\sqrt{2}t^2}{16\sigma^2 w }\right)
\end{aligned}
\end{equation}
\end{lemma}

\textbf{Proof}. Note that $\mathbf{X}_{k}$ is a self-adjoint matrix, therefore $\mathbf{X}_{k}$ is always diagonalizable with real eigenvalues. Denote $\mathbf{X}_{k} = \mathbf{Q}_{k}^* \mathbf{D}_{k} \mathbf{Q}_{k}$, where $\mathbf{Q}_{k}$ is a unitary matrix and $\mathbf{D}_{k}= \text{diag}(\eta^{(1)}_k, \eta^{(2)}_k, \cdots, \eta^{(d)}_k)$ is a diagonal matrix with $\eta^{(i)}_k \in \mathbb{R}$. For any $\sigma^2$-sub-Gaussian noise $\epsilon_k$,  we have the following bound on the moment generating function 

\begin{equation}
\begin{aligned}
    \mathbb{E}_{k-1} \left[\text{exp} \left({\theta \mathbf{X}_{k}\epsilon_k} \right) \right] 
    &= \sum_{n=0}^\infty  \frac{\theta^n \mathbf{X}_{k}^n \mathbb{E}_{k-1}[\epsilon_k^n] }{n!} \\
    &= \sum_{n=0}^\infty  \frac{\theta^n \mathbf{Q}_{k}^* \mathbf{D}_{k}^n \mathbf{Q}_{k} \mathbb{E}_{k-1}[\epsilon_k^n] }{n!} \\
    &= \mathbf{Q}_{k}^* \left( \sum_{q=0}^\infty  \frac{\theta^{2q}  \mathbf{D}_{k}^{2q}   \mathbb{E}_{k-1}[\epsilon_k^{2q}] }{(2q)!} + \sum_{q=0}^\infty  \frac{\theta^{2q+1} \mathbf{D}_{k}^{2q+1}  \mathbb{E}_{k-1}[\epsilon_k^{2q+1}] }{(2q+1)!} \right) \mathbf{Q}_{k} \\
    &\preceq \mathbf{I} +  \frac{7}{6}\theta^{2} \mathbf{X}_{k}^{2}  (2\sigma^2) +  \mathbf{Q}_{k}^* \left( \sum_{q=2}^\infty     \frac{4q^2 + 6q + 3}{(2q+1)!}  \theta^{2q} \mathbf{D}_{k}^{2q}  (2\sigma^2)^q q! \right)\mathbf{Q}_{k}\\
    &\preceq  \left( \sum_{q=0}^\infty \frac{2^q \theta^{2q} \mathbf{X}_{k}^{2q}  (2\sigma^2)^q }{q!}  \right)=  \text{exp} \left(2\sqrt{2}\sigma^2 \theta^2 \mathbf{X}^2_k  \right) \\
\end{aligned}
\end{equation}

where the last inequality is because of the algebraic properties $(4q^2+6q+3)/(2q+1) \leq 2^q$ when $q>3$, and $[(4q^2+6q+3)q!q!]/(2q+1)! \leq 2^q$ when $q=2$ and $q=3$. Besides, the first inequality is from Lemma \ref{lem: subgaussian_moment_bound} and thus the following fact 

\begin{equation}
\begin{aligned}
  &  \sum_{q=1}^\infty  \frac{\theta^{2q+1} \left(\eta^{(i)}_{k}\right)^{2q+1}  \mathbb{E}_{k-1}[\epsilon_k^{2q+1}] }{(2q+1)!} \\
  &\leq \sum_{q=1}^\infty  \frac{\theta^{2q} \left(\eta^{(i)}_{k}\right)^{2q}  \mathbb{E}_{k-1}[\epsilon_k^{2q}] +  \theta^{2q+2} \left(\eta^{(i)}_{k}\right)^{2q+2}  \mathbb{E}_{k-1}[\epsilon_k^{2q+2}] }{2(2q+1)!} \\
  &\leq  \sum_{q=1}^\infty     \frac{ \theta^{2q} \left(\eta^{(i)}_{k}\right)^{2q} (2\sigma^2)^q q! +   \theta^{2q+2} \left(\eta^{(i)}_{k}\right)^{2q+2}  (2\sigma^2)^{q+1} (q+1)!  }{(2q+1)!}  \\
  &\leq  \frac{1}{6}  \theta^{2} \left(\eta^{(i)}_{k}\right)^{2q} (2\sigma^2)^q  + \sum_{q=2}^\infty     \frac{4q^2 + 2q +1}{(2q+1)!}  \theta^{2q} \left(\eta^{(i)}_{k}\right)^{2q} (2\sigma^2)^q q!  \\
\end{aligned}
\end{equation}

Therefore, by the master tail bound (Lemma \ref{lem: master_tail_bound}), we have the following inequality

\begin{equation}
\begin{aligned}
\mathbb{P}\left (\exists k \geq 0: \lambda_{\max }\left(\sum_{i=1}^k \mathbf{X}_i \epsilon_i \right) \geq t \text { and } \lambda_{\max }\left(\sum_{i=1}^k \mathbf{X}_i^2 \right) \leq w\right ) \leq d \cdot \inf _{\theta>0} \mathrm{e}^{-\theta t+ 2\sqrt{2}\sigma^2 \theta^2 w}
\end{aligned}
\end{equation}

Note that the quadratic function $f(\theta) = -\theta t+ 2\sqrt{2}\sigma^2 \theta^2 w $ is minimized by $\theta = t/(4\sqrt{2}\sigma^2 w)$. Therefore the infimum is $\inf_{\theta > 0} f(\theta) = -\frac{t^2}{8\sqrt{2}\sigma^2 w}$ and we arrive to the conclusion in the lemma. \hfill $\square$

\subsection{Auxiliary Definitions and Lemmas}

\begin{definition}
\label{def: subgaussian}
\textbf{\upshape (Sub-Gaussian Random Variables)} A random variable $X \in \mathbb{R}$ is said to be sub-Gaussian with mean $\mu$ and variance proxy $\sigma^{2}$ if 
\begin{equation}
\mathbb{P} \left(|X-\mu| \geq t\right) \leq 2 \exp\left (-\frac{t^2}{2\sigma^2} \right)
\end{equation} 
\end{definition}

\begin{definition}
\label{def: adapted_and_previsible_matrices} \textbf{\upshape (Adapted Sequence and Previsible Sequence)}
Let $(\Omega,  \mathcal{F}, \mathbb{P}) $ be a probability space and let $\mathcal{F}_0 \subseteq \mathcal{F}_1 \subseteq \mathcal{F}_2 \subseteq \cdots \subseteq \mathcal{F} $ be a filtration of the master sigma-algebra. We say a sequence of random matrices $\{\mathbf{X}_k\}$ is \textbf{adapted} to the filtration when each $\mathbf{X}_k$ is measurable with respect to $\mathcal{F}_k$. We say a sequence of random matrices $\{\mathbf{V}_k\}$ is \textbf{previsible} to the filtration when each $\mathbf{V}_k$ is measurable with respect to $\mathcal{F}_{k-1}$.
\end{definition}

\begin{lemma}
\label{lem: master_tail_bound}
\textbf{\upshape (Master Tail Bound for Adapted Sequences, Theorem 2.3 in \citet{Tropp2011freedmans})}. Consider an adapted sequence $\left\{\mathbf{X}_{k}\right\}$ and a previsible sequence $\left\{\mathbf{V}_{k}\right\}$ of self-adjoint matrices with dimension $d \times d$. Assume these sequences satisfy the relations

\begin{equation}
\log \mathbb{E}_{k-1} \mathrm{e}^{\theta \mathbf{X}_{k}} \preceq g(\theta) \cdot \mathbf{V}_{k} \quad \text { almost surely for each } \theta>0
\end{equation}

where the function $g:(0, \infty) \rightarrow[0, \infty]$. Define the partial sum processes

\begin{equation}
\mathbf{Y}_{k}:=\sum_{j=1}^{k} \mathbf{X}_{j} \quad \text {and} \quad \mathbf{W}_{k}:=\sum_{j=1}^{k} \mathbf{V}_{j}
\end{equation}

Then, for all $t, w \in \mathbb{R}$

\begin{equation}
\mathbb{P}\left (\exists k \geq 0: \lambda_{\max }\left(\mathbf{Y}_{k}\right) \geq t \text { and } \lambda_{\max }\left(\mathbf{W}_{k}\right) \leq w\right ) \leq d \cdot \inf _{\theta>0} \mathrm{e}^{-\theta t+g(\theta) \cdot w}
\end{equation}
\end{lemma}

\begin{lemma} \textbf{\upshape (Property of the Dilation Operation)}
\label{lem: property_of_dilation} For any two matrices $\mathbf{X}, \mathbf{Y} \in \mathbb{R}^{d_1\times d_2}$. define the dilation matrix $\mathbf{Z}$ to be

\begin{equation}
\mathbf{Z} = 
\begin{bmatrix}
& \mathbf{0} & \mathbf{X} \\
& \mathbf{Y}^T  & \mathbf{0}
\end{bmatrix}
\end{equation}

Then we have the property that $\vertiii{\mathbf{Z}}_{op} = \max \left \{ \vertiii{\mathbf{X}}_{op}, \vertiii{\mathbf{Y}}_{op} \right \}$. 

\end{lemma}
\textbf{Proof}. Note that for any vector $w^T = [u^T, v^T] \in \mathbb{R}^{d_1+d_2}$ with $u \in \mathbb{R}^{d_2}$ and $v \in \mathbb{R}^{d_1}$, we have

\begin{equation}
\begin{aligned}
     \left  \|\mathbf{Z}w \right\|_2 &= \left \| \begin{bmatrix}
    \mathbf{X}v\\
    \mathbf{Y}^Tu
    \end{bmatrix} \right \|_2 = \sqrt{\left \|\mathbf{X}v \right\|_2^2 + \left \|\mathbf{Y}^T u \right \|_2^2} \\
    &\leq \sqrt{\vertiii{\mathbf{X}}_{op}^2 \left \|v \right\|_2^2 + \vertiii{\mathbf{Y}}_{op}^2 \left \| u \right \|_2^2} \\
    &\leq \max \left \{ \vertiii{\mathbf{X}}_{op}, \vertiii{\mathbf{Y}}_{op} \right \} \sqrt{ \left \|v \right\|_2^2 + \left \| u \right \|_2^2} \\
    &\leq  \max \left \{ \vertiii{\mathbf{X}}_{op}, \vertiii{\mathbf{Y}}_{op} \right \} \|w\|_2
\end{aligned}
\end{equation}

Therefore we have the conclusion that $ \vertiii{\mathbf{Z}}_{op} \leq \max \left \{ \vertiii{\mathbf{X}}_{op}, \vertiii{\mathbf{Y}}_{op} \right \} $. Moreover, without loss of generality assume that $  \vertiii{\mathbf{X}}_{op} \geq \vertiii{\mathbf{Y}}_{op} $, then take $w^T = [0, v_0^T]$, where $\|\mathbf{X} v_0\|_2 = \vertiii{X}_{op}\|v_0\|_2$ (by definition of the operator norm, such $v_0$ always exists), we have

\begin{equation}
\begin{aligned}
     \left  \|\mathbf{Z}w \right\|_2 = \|\mathbf{X} v_0\|_2 = \vertiii{\mathbf{X}}_{op} \|v_0\|_2 = \vertiii{\mathbf{X}}_{op} \|w\|_2 
\end{aligned}
\end{equation}

Therefore $\vertiii{\mathbf{Z}}_{op} = \vertiii{\mathbf{X}}_{op}$ in this case. We arrive to the conclusion that $\vertiii{\mathbf{Z}}_{op} = \max \left \{ \vertiii{\mathbf{X}}_{op}, \vertiii{\mathbf{Y}}_{op} \right \}$. \hfill $\square$

\begin{lemma}
\label{lem: subgaussian_moment_bound}
\textbf{\upshape (Moment Bounds for Sub-Gaussian Random Varibales)}For any zero-mean $\sigma^2$-sub-Gaussian random variable $Z$ and any integer $q \in \mathbb{N}$, we have the following bounds on the moments of $Z$.

\begin{equation}
\begin{aligned}
    \mathbb{E}\left[Z^{2q} \right] \leq 2  (2\sigma^2)^q q!, \quad \mathbb{E}\left[Z^{2q+1} \right] \leq (2\sigma^2)^q q! +  (2\sigma^2)^{q+1} (q+1)!
\end{aligned}
\end{equation}

\end{lemma}
\textbf{Proof.}
For any even number $2q$ with $q \in \mathbb{N}$, we have the following bound on the $2q$-th moment of the sub-Gaussian random variable.

\begin{equation}
\begin{aligned}
    \mathbb{E}\left[Z^{2q} \right] 
    &= \int_{0}^{+\infty}  \mathbb{P}(Z^{2q} \geq u )du = \int_{0}^{+\infty}  \mathbb{P}(|Z|\geq u^{1/(2q)} )du \\
    &\leq  2 \int_{0}^{+\infty} \text{exp}( - \frac{u^{1/q}}{2\sigma^2} )du \\
    &= 2  (2\sigma^2)^q q \int_{0}^{+\infty} \text{exp}( - v )v^{q-1} dv\\
    &=  2  (2\sigma^2)^q q!
\end{aligned}
\end{equation}

where the third equality is by change of variable $ u = (2\sigma^2 v)^q$ and thus $du =  (2\sigma^2)^q q v^{q-1} dv $. The last equality is because $\Gamma(q) = (q-1)!$. Now for any odd number $2q+1$, we have the following

\begin{equation}
\tag*{$\square$}
\begin{aligned}
    \mathbb{E}\left[Z^{2q+1} \right] 
    &= \mathbb{E}\left[Z^{q} Z^{q+1} \right]\leq \mathbb{E}\left[\frac{Z^{2q} + Z^{2q+2}}{2} \right] \leq (2\sigma^2)^q q! +  (2\sigma^2)^{q+1} (q+1)! 
\end{aligned}
\end{equation}

\section{Always Valid Inequality for Sum of Matrices and  Sub-Exponentials}
\label{app:subExponential}

In this section, we present the always-valid inequality for the sum of products of matrices and sub-Exponential random variables, which may also be of independent interest to the readers. We prove such inequalities for both self-adjoint matrices and general rectangular matrices. Similarly, the latter inequality is based on the former one. The proof here provides an alternative way to establish the always-valid inequality apart from the approach in Appendix \ref{app:subGaussian}, which does not rely on Lemma \ref{lem: master_tail_bound}, but the proof is still based on the idea behind Lieb's Lemma.

In fact, Lemma \ref{lem: master_tail_bound} cannot be easily applied for sub-Exponential random variables because the upper bound function $g(\theta)$ cannot be easily established for all $\theta > 0$. However, in this section we show that $g(\theta)$ can be constructed for $\theta$ in a restrained region (Lemma \ref{lem: sub_exponential_log_exponential_bound}). Note that here we also require the matrices to have bounded operator norm, which is not hard to satisfy because operator norm is always bounded by the Frobenius norm and the requirement can be achieved through normalization. Such an additional requirement is needed due to the uniqueness of the Hoeffding-type tail of sub-Exponential random variables, and more specifically, it is used in the proof of Lemma \ref{lem: sub_exponential_log_exponential_bound}.

\subsection{Proof of Lemma \ref{lem: always_valid_inequality_rectangular_subexponential}}

\begin{lemma}
\label{lem: always_valid_inequality_rectangular_subexponential}  \textbf{\upshape (Always Valid Inequality for Rectangular Matrices, Sub-Exponential Case)}
For any sequence of self-adjoint matrices $\{\mathbf{X}_i\}_{i=1}^{\infty}$ with $\mathbf{X}_i \in \mathbb{R}^{d\times d}$ and satisfies $\vertiii{\mathbf{X}_i}_{op} \leq x_{\max}$, for any sequence of conditional $\lambda$-sub-Exponential noises $\{\epsilon_i\}_{i=1}^{\infty}$, define the following quantity 

\begin{equation}
S_k = \max \left \{\vertiii{\sum_{i=1}^k \mathbf{X}_i  \mathbf{X}_i^T}_{op}, \vertiii{\sum_{i=1}^k \mathbf{X}_i^T \mathbf{X}_i}_{op}\right \}
\end{equation}

Then we have that for any $t > 0$ and $w>0$,

\begin{equation}
\mathbb{P}\left (\exists k \geq 0: \lambda_{\max }\left(\sum_{i=1}^k \mathbf{X}_i \epsilon_i \right) \geq t \text { and } S_k \leq w\right ) \leq \left \{
\begin{aligned}
 & (d_1 + d_2) \cdot  \mathrm{exp} \left(- \frac{t^2}{576\lambda^2 w }\right) &\text{ if $t\leq24\lambda w/x_{\max}$} \\
 & (d_1 + d_2) \cdot  \mathrm{exp} \left(- \frac{t}{24\lambda x_{\max}}\right) & \text{ if $t\geq24\lambda w/x_{\max}$}
\end{aligned}
\right.
\end{equation}

\end{lemma}

\textbf{Proof}. The proof here follows the same idea as the proof of Lemma \ref{lem: always_valid_inequality_rectangular_subgaussian}, where we utilize the dilation trick to expand the rectangular matrix into a self-adjoint matrix, i.e., we can define the following matrix for each $\mathbf{X}_i$ and apply Lemma \ref{lem: always_valid_inequality_self_adjoint_subexponential} to it.

\begin{equation}
\mathbf{Z}_i = 
\begin{bmatrix}
& \mathbf{0} & \mathbf{X}_i \\
& \mathbf{X}_i^T  & \mathbf{0}
\end{bmatrix}
\end{equation}

The rest of the proof follows by using Lemma \ref{lem: property_of_dilation}. Again, we emphasize that the upper bound $x_{\max}$ of $\vertiii{\mathbf{X}_i}_{op}$ is needed in this Lemma because in Lemma \ref{lem: sub_exponential_log_exponential_bound}, we need it to bound the tail behavior of sub-exponentials. \hfill $\square$

\begin{lemma}
\label{lem: always_valid_inequality_self_adjoint_subexponential}
\textbf{\upshape (Always Valid Inequality for Bounded Self-adjoint Matrices, Sub-Exponential Case)}
For any sequence of self-adjoint matrices $\{\mathbf{X}_i\}_{i=1}^{\infty}$ with $\mathbf{X}_i \in \mathbb{R}^{d\times d}$ and satisfies $\vertiii{\mathbf{X}_i}_{op} \leq x_{\max}$, for any sequence of conditional $\lambda$-sub-Exponential noises $\{\epsilon_i\}_{i=1}^{\infty}$, we have that for any $t > 0$ and $w>0$,

\begin{equation}
\mathbb{P}\left (\exists k \geq 0: \lambda_{\max }\left(\sum_{i=1}^k \mathbf{X}_i \epsilon_i \right) \geq t \text { and } \lambda_{\max }\left(\sum_{i=1}^k \mathbf{X}_i^2 \right) \leq w\right ) \leq \left \{
\begin{aligned}
 &d \cdot  \mathrm{exp} \left(- \frac{t^2}{576\lambda^2 w }\right) &\text{ if $t\leq24\lambda w /x_{\max}$} \\
 &d \cdot  \mathrm{exp} \left(- \frac{t}{24\lambda x_{\max}}\right) & \text{ if $t\geq24\lambda w /x_{\max}$}
\end{aligned}
\right.
\end{equation}

\end{lemma}

\textbf{Proof}. The upper bound of $\vertiii{\mathbf{X}_i}_{op}$ $x_{\max}$ is needed in this Lemma because in Lemma \ref{lem: sub_exponential_log_exponential_bound}, we need it to bound the tail behavior of sub-exponentials. The proof of this lemma follows the idea of Theorem 3.1 in \citet{Tropp2011freedmans}, but the difference is that we need to establish the results for $\theta$ in a constrained region (the region in Lemma \ref{lem: sub_exponential_log_exponential_bound}) instead of on $\mathbb{R}_+$, and thus optimize the lower bound in Lemma \ref{lem: lower_bound_of_G} over the constrained region. That is also the reason why we observe two types of tail bounds (Hoeffding and Bernstein) when $t$ is in different regions in the equation above.

Define the following two sequences of partial sums $\{\mathbf{Y}_k, \mathbf{Z}_k\}_{k=0, 1, 2 \cdots}$

\begin{equation}
\begin{aligned}
       &\mathbf{Y}_0 := 0 \text{ and }   &\mathbf{Y}_k := \sum_{i=1}^k \mathbf{X}_i \\
       &\mathbf{W}_0 := 0 \text{ and }   &\mathbf{W}_k := \sum_{i=1}^k \mathbf{X}_i^2  \\
\end{aligned}
\end{equation}

Now let $g(\theta) = 144\lambda^2\theta^2$ for $\theta \in (0, \frac{1}{12\lambda x_{\max}}]$ as in the right hand side of the equation in Lemma \ref{lem: sub_exponential_log_exponential_bound}. Define the function $G_\theta(\mathbf{Y}, \mathbf{W})$ and the real-valued random process $S_k$ to be:

\begin{equation}
\begin{aligned}
\label{eqn: function_G_theta}
    &G_\theta(\mathbf{Y}, \mathbf{W}):= \text{tr}\left(\exp \left( \theta \mathbf{Y} - g(\theta) \mathbf{W}\right) \right) \\
    &S_k := S_k(\theta) = G(\theta)(\mathbf{Y}_k, \mathbf{W}_k)
\end{aligned}
\end{equation}

Define a stopping time $\kappa$ that is the first time when the partial sum $\mathbf{Y}_k$ is larger than $t$ but the other partial sum $\mathbf{W}_k$ is bounded above by $w$, i.e.,

\begin{equation}
\begin{aligned}
    &\kappa :=  \inf \{ {k \in \mathbb{N}}: \lambda_{\max}(\mathbf{Y}_k) \geq t \text{ and }  \lambda_{\max}(\mathbf{W}_k) \leq w )\}
\end{aligned}
\end{equation}

When the infimum is empty, the stopping time is defined to be $\kappa = \infty$. Now by Lemma \ref{lem: sub_exponential_log_exponential_bound}, Lemma \ref{lem: super-martingale}, and Lemma \ref{lem: lower_bound_of_G}, we know that for any $\theta \in (0, \frac{1}{12\lambda x_{\max} }]$, we have

\begin{equation}
\begin{aligned}
    d \geq \mathbb{E}[S_k] \geq \mathbb{E}[S_\kappa \mid \kappa < \infty] \geq \mathbb{P}(\kappa < \infty) \cdot \inf S_\kappa \geq  \mathbb{P}(\kappa < \infty) \cdot \exp(\theta t - g(\theta) w)
\end{aligned}
\end{equation}

Therefore we get

\begin{equation}
\begin{aligned}
 \mathbb{P}(\kappa < \infty) = \mathbb{P}\left (\exists k \geq 0: \lambda_{\max }\left(\sum_{i=1}^k \mathbf{X}_i \epsilon_i \right) \geq t \text { and } \lambda_{\max }\left(\sum_{i=1}^k \mathbf{X}_i^2 \right) \leq w\right ) \leq  \exp( - \theta t + g(\theta) w)
\end{aligned}
\end{equation}

Now it remains to optimize the right-hand side on the given range of $\theta$, note that the function

\begin{equation}
\begin{aligned}
 - \theta t + g(\theta) w =  - \theta t + 144\lambda^2\theta^2 w = 144\lambda^2 w (\theta - \frac{t}{288\lambda^2 w})^2 - \frac{t^2 }{576 \lambda^2 w}
\end{aligned}
\end{equation}

Therefore when $t \geq (24\lambda w/x_{\max})$,  the function is minimized at the boundary point $1/(12\lambda x_{\max})$, and the function value is 
\begin{equation}
\frac{w}{x_{\max}^2} - \frac{t}{12\lambda x_{\max}} \leq - \frac{t}{24\lambda x_{\max}}
\end{equation}
When $t\leq (24\lambda w/x_{\max})$, the function is minimized at the lowest point of the parabola. Therefore we arrive at the conclusions in the Lemma. \hfill $\square$

\begin{lemma}
\label{lem: sub_exponential_log_exponential_bound}
For any sequence of self-adjoint matrices $\{\mathbf{X}_i\}_{i=1}^{\infty}$ with $\mathbf{X}_i \in \mathbb{R}^{d\times d}$ that satisfies $\vertiii{\mathbf{X}_i}_{op} \leq x_{\max}, \forall i$ for some $x_{\max} \geq 0$,  and any sequence of conditional $\lambda$-sub-Exponential noises $\{\epsilon_i\}_{i=1}^{\infty}$, we have that for any $\theta \in (0, \frac{1}{12 \lambda x_{\max}}]$, we have
\begin{equation}
\begin{aligned}
       \log \mathbb{E}_{k-1} \left[ \exp(\theta \mathbf{X}_k \epsilon_k ) \right] 
        \preceq  144 \lambda^2 \theta^2 \mathbf{X}_k^2 
\end{aligned}
\end{equation}
\end{lemma}
\textbf{Proof.}
Assume for now that $\vertiii{\mathbf{X}_k}_{op} > 0$ since if it is zero, it implies the matrix is a zero matrix and the expectation  $ \mathbb{E}_{k-1} \left[ \exp(\theta \mathbf{X}_k \epsilon_k ) \right] $ is 1 and thus the inequality in the lemma holds trivially. For any $k \in [T]$, by Taylor expansion and the fact that each $\epsilon_k$ is zero-mean, we have that

\begin{equation}
\begin{aligned}
       \mathbb{E}_{k-1} \left[ \exp(\theta \mathbf{X}_k \epsilon_k ) \right] &=  \mathbb{E}_{k-1} \left[\mathbf{I} + \theta^2 \mathbf{X}_k^2 \epsilon_k^2 \left ( \frac{1}{2!} +   \frac{\theta \mathbf{X}_k \epsilon_k }{3!} +  \frac{\theta^2 \mathbf{X}_k^2 \epsilon_k^2 }{4!} + \cdots\right) \right]\\
       &\preceq \mathbb{E}_{k-1} \left[\mathbf{I} + \theta^2 \mathbf{X}_k^2 \epsilon_k^2 \left ( \frac{1}{2!} +   \frac{\theta \vertiii{\mathbf{X}_k}_{op} |\epsilon_k| }{3!} +  \frac{ \theta^2 \vertiii{ \mathbf{X}_k}_{op}^2 |\epsilon_k|^2 }{4!} + \cdots\right) \right] \\
       &=  \mathbf{I} + \theta^2 \mathbf{X}_k^2 \mathbb{E}_{k-1} \left[ \epsilon_k^2 \left ( \frac{1}{2!} +   \frac{\theta \vertiii{\mathbf{X}_k}_{op} |\epsilon_k| }{3!} +  \frac{ \theta^2  \vertiii{ \mathbf{X}_k}_{op}^2 |\epsilon_k|^2 }{4!} + \cdots\right) \right]\\
       &\preceq \exp \left ( \theta^2 \mathbf{X}_k^2 \mathbb{E}_{k-1} \left[ \epsilon_k^2 \frac{ \psi \left( \theta \vertiii{\mathbf{X}_k}_{op} |\epsilon_k|\right ) - \theta \vertiii{\mathbf{X}_k}_{op} |\epsilon_k|  }{\theta^2 \vertiii{\mathbf{X}_k}_{op}^2 |\epsilon_k|^2}  \right] \right)
\end{aligned}
\end{equation}

where the first inequality is by the fact that for any self-adjoint matrix $\mathbf{A}$ and any natural number $p \geq 2$, $\mathbf{A}^p  \preceq  \mathbf{A}^2 \vertiii{\mathbf{A}^{p-2}}_{op}$ as the matrices $\mathbf{A}^{2}$ and $(\vertiii{\mathbf{A}^{p-2}}_{op} \mathbf{I} - \mathbf{A}^{p-2})$ are positive definite by definition of the operator norm. The last inequality is because $\mathbf{X}_k^2$ and its powers are always positive definite. The function $\psi$ is the Orlicz-1 norm function defined in Definition \ref{def: orlicz_1_norm}. Now since logarithm is an operator monotone function for matrices \citep{Uchiyama2010Operator}, we have 

\begin{equation}
\begin{aligned}
       \log \mathbb{E}_{k-1} \left[ \exp(\theta \mathbf{X}_k \epsilon_k ) \right] 
       &\preceq  \theta^2 \mathbf{X}_k^2 \mathbb{E}_{k-1} \left[ \epsilon_k^2 \frac{ \psi \left( \theta \vertiii{\mathbf{X}_k}_{op} |\epsilon_k|\right ) - \theta \vertiii{\mathbf{X}_k}_{op} |\epsilon_k|  }{\theta^2 \vertiii{\mathbf{X}_k}_{op}^2 |\epsilon_k|^2}   \right] \\
       & \preceq  144 \lambda^2 \theta^2 \mathbf{X}_k^2 \mathbb{E}_{k-1} \left[ \psi \left( \frac{|\epsilon_k|}{12\lambda} \right ) \right]  \\
       & \preceq  144 \lambda^2 \theta^2 \mathbf{X}_k^2 
\end{aligned}
\end{equation}

where the second inequality is because $(\psi(x) -x)/x^2 $ is a monotonically increasing function and that $\frac{1}{12\lambda} \geq \theta \vertiii{\mathbf{X}_k}_{op} > 0$ when $0 < \theta \leq \frac{1}{12\lambda x_{\max}}$. The third inequality is by Lemma \ref{lem: subexponential_orlicz_1_norm} and the definition of Orlicz-1 norm in Definition \ref{def: orlicz_1_norm}. We arrive at the conclusions in the lemma. \hfill $\square$

\subsection{Auxiliary Definitions and Lemmas}

\begin{definition}
\label{def: subexponential}
\textbf{\upshape (Sub-Exponential Random Variables)} A zero-mean random variable $X \in \mathbb{R}$ is said to be sub-Exponential with parameter $\lambda$ if  $\mathbb{E}(X) = 0$ and for all $s \leq \frac{1}{\lambda}$, we have
\begin{equation}
\mathbb{E}(\exp(sX)) \leq \exp({s^2\lambda^2})
\end{equation} 
\end{definition}

\begin{definition}
\label{def: orlicz_1_norm}
\textbf{\upshape (Definition of Orlicz-1 Norm)} Define $\psi: \mathbb{R}^+ \rightarrow \mathbb{R}^+$ be the Orlicz-1 function, i.e., $ \psi(x) = e^x - 1$. Then the Orlicz-1 norm of a random variable $X$ is defined as

\begin{equation}
    \|X\|_\psi = \inf \{ C>0 \mid \mathbb{E} [\psi(\frac{|X|}{C})] \leq 1\}
\end{equation}
\end{definition}

\begin{lemma}
\label{lem: subexponential_tail}
\textbf{\upshape (Sub-Exponential Tails)} For a zero-mean sub-Exponential random variables $X$ with parameter $\lambda$, we have that for any $t>0$,
\begin{equation}
\begin{aligned}
    \mathbb{P} \left(|X| \geq t \right) \leq 2\exp( - \frac{t}{4\lambda}) 
\end{aligned}
\end{equation}
\end{lemma}
\textbf{Proof}. Note that by Markov's inequality, we have for $X > 0$ and any $ 0\leq s \leq \frac{1}{\lambda}$
\begin{equation}
\begin{aligned}
    \mathbb{P} \left(X \geq t \right) \leq  \mathbb{E}[\exp({sX- st})] \leq \exp({s^2\lambda^2 - st})
\end{aligned}
\end{equation}

Now we optimize over the choice of $s$ on the interval. Since the function $\phi(s) = \lambda^2 s^2 - ts$ is minimized at $s = {t}/{2\lambda^2}$, we get the optimal value $\phi(s) = -\frac{t^2}{4\lambda^2}$ at $s = {t}/{2\lambda^2}$ when $t \leq 2\lambda$ and $\phi(s) = 1 - \frac{t}{\lambda} \leq - \frac{t}{2\lambda}$ otherwise. Therefore  we have

\begin{equation}
\mathbb{P} \left(X \geq t \right) \leq     \left \{
\begin{aligned}
    &\exp(-\frac{t^2}{4\lambda^2}) \quad &\text{when $t \leq 2\lambda$},\\
    &\exp( - \frac{t}{2\lambda}) \quad &\text{when $t \geq 2\lambda$},\\
\end{aligned}
    \right. 
\end{equation}

Therefore by symmetry and the union bound we have

\begin{equation}
\mathbb{P} \left(|X| \geq t \right) \leq     \left \{
\begin{aligned}
    &2\exp(-\frac{t^2}{4\lambda^2}) \quad &\text{when $t \leq 2\lambda$},\\
    &2\exp( - \frac{t}{2\lambda}) \quad &\text{when $t \geq 2\lambda$}\\
\end{aligned}
    \right. 
\end{equation}

Now the above shows Bernstein behavior when $t\leq 2\lambda$ and Hoeffding behavior when $t\geq 2\lambda$, which is another important characterization of sub-Exponential random variables. Also note that when $t \leq 2\lambda$, we have 

\begin{equation}
    2\exp(- \frac{t}{4\lambda}) \geq 2 \exp (-\frac{1}{2}) \geq 1
\end{equation}

Therefore in general we have for any $t > 0$

\begin{equation}
\tag*{$\square$}
\begin{aligned}
    \mathbb{P} \left(|X| \geq t \right) \leq 2\exp( - \frac{t}{4\lambda}) 
\end{aligned}
\end{equation}

\begin{lemma}
\label{lem: subexponential_orlicz_1_norm}
\textbf{\upshape (Sub-Exponentials Have Bounded Orlicz-1 Norm)} For a zero-mean sub-Exponential random variables $X$ with parameter $\lambda$, we have  
\begin{equation}
\begin{aligned}
    \|X\|_\psi \leq 12 \lambda
\end{aligned}
\end{equation}
\end{lemma}
\textbf{Proof.} Although this lemma is a quite well-known fact and it is also used as one of the definitions of sub-Exponential random variables (i.e., the Orlicz-1 norm is finite), we provide the proof here for completeness. The bound in the lemma may not be tight but it is enough for our arguments in this paper. Note that we have the following bound from the definition of sub-Exponential random variables

\begin{equation}
\begin{aligned}
       \mathbb{E} [\psi(\frac{|X|}{C})] &=  \mathbb{E} [\text{exp}(\frac{|X|}{C}) - 1] \\
       &= \int_{0}^{\infty} \mathbb{P} \left( \text{exp}(\frac{|X|}{C}) \geq t \right) dt - 1 \\
       &= \int_{-\infty}^{\infty} \mathbb{P} \left( \frac{|X|}{C} \geq m \right) e^m dm - 1 \\
       &=  \int_{0}^{\infty} \mathbb{P} \left( \frac{|X|}{C} \geq m \right) e^m dm  \\
       &\leq  \int_{0}^{\infty} 2 \exp\left( (1-\frac{C}{4\lambda}) m \right) dm  \\
       & = -  \frac{8\lambda}{4\lambda - C} \qquad \text{ for } C > 4\lambda 
\end{aligned}
\end{equation}

where the second equality is by change of variable $t = e^m$ and the third equality is because $\int_{-\infty}^0 e^m dm = 1$. The first inequality is by Lemma \ref{lem: subexponential_tail}. Now for the above result to be less than $1$, we need $C\geq 12 \lambda$. By the definition of Orlicz-1 norm (Definition \ref{def: orlicz_1_norm}), we have $\|X\|_\psi \leq 12\lambda$. \hfill $\square$

\begin{lemma} \textbf{(\upshape Super-martingale, Lemma 2.1 in \citet{Tropp2011freedmans})}
\label{lem: super-martingale}
For each fixed $\theta > 0$, the random process $\{S_k(\theta): k=0,1,2, \cdots\}$ defined in Eqn. (\ref{eqn: function_G_theta}) is a positive super-martingale whose initial value $S_0 = d$
\end{lemma}

\begin{lemma} \textbf{(\upshape Lower Bound of $G_{\theta}$, Lemma 2.2 in \citet{Tropp2011freedmans})}
\label{lem: lower_bound_of_G}
Suppose that $\lambda_{\max}(\mathbf{Y}) \geq t$ and that $\lambda_{\min}(\mathbf{W}) \leq w$. For any $\theta> 0$, the $G_{\theta}(\mathbf{Y}, \mathbf{W})$ defined in Eqn. (\ref{eqn: function_G_theta}) satisfies
\begin{equation}
\begin{aligned}
        G_\theta(\mathbf{Y}, \mathbf{W}) \geq \exp(\theta t - g(\theta) w)
\end{aligned}
\end{equation}
\end{lemma}

\section{Experiment Details}
\label{app:experiments}

We provide more details for the experimental results in Section \ref{sec: experiments} in this section. We use the PyTorch package for optimizing the nuclear-norm regularized negative log likelihood loss function, which provides automatic differentiation and back-propagation. However, the singular value decomposition (SVD) package of PyTorch is quite sensitive to numerical values, and some parameter tuning is necessary during the experiments.

We generate the target matrices by first generating random vectors, where we uniformly randomly generate each element of the vector from $(0, 1)$ and enlarge them by a constant $c$. For Gaussian we choose $c = 10$; for Binomial and Poisson we choose $c = 1$ because setting large $c$ will result in computational problems: their $G'(x)$ either make the labels $y_i$ too close so that Frobenius norm will converge very slowly (Binomial) or too large to optimize and the computational process fail (Poisson). Then for each row of the target matrix, we choose from the generated random vectors randomly so that the matrix is low-rank. The indices chosen at each iteration are generated uniformly randomly from all the indices available. After the indices are chosen, the label is generated by passing the chosen matrix value through the link function $G'(x)$ and adding the corresponding noise.

For the optimization algorithm, we use gradient descent with no weight decay, momentum or Nesterov momentum. We have tuned the learning rate by grid-search in $\{0.001, 0.005, 0.01, 0.05, 0.1, 0.3, 1, 2, 5\}$ and we find that setting the learning rate to $1$ or $2$ will generate the best performance. Setting larger or smaller learning rate will possibly generate slow convergence or make the SVD package fail.

Some other hyper-parameters we have used:

\begin{itemize}
  \setlength\itemsep{1em}
    \item For the confidence level $\alpha$, we set $\alpha = 0.01$.
    \item For the rank $r$, we set $r = 1$ for the $5 \times 5$ matrix and $r = 2$ for the $10 \times 10$ matrix.
    \item For the parameter $S_t$, we compute it at every epoch based on the operator norm package from PyTorch.
    \item For the regularization parameter $\lambda_t$, we set it exactly the same as our theoretical parts.
    \item For the number of iterations in each epoch for the optimization of the loss function, we set it to be 100 and find it to be sufficient.
\end{itemize}

\end{document}

%% file: macro&setup/myPKG.tex
\usepackage{amsmath,amsthm,amssymb,amsfonts}
\usepackage[utf8x]{inputenc}
\usepackage{mathrsfs}
\usepackage{verbatim}
\usepackage{multirow}
\usepackage{longtable}
\usepackage{enumerate}
\usepackage{color}
\usepackage{tikz}

\usepackage{subfigure}
\usepackage[english]{babel}
\usepackage{graphicx}
\usepackage{framed}
\usepackage[normalem]{ulem}
\usepackage{indentfirst}
\usepackage{bbm}

\usepackage[italicdiff]{physics}
\usepackage[T1]{fontenc}
\usepackage{wrapfig}
\usepackage{lmodern,mathrsfs}
\usepackage{multicol}

\theoremstyle{definition}
\newtheorem{theorem}{Theorem}[section]
\newtheorem{lemma}{Lemma}[section]
\newtheorem{definition}{Definition}[section]
\newtheorem{corollary}[theorem]{Corollary}

\newtheorem{remark}{Remark}

%% file: macro&setup/myDeco.tex
\newcommand{\mydeco}[2]{\begin{#1}#2\end{#1}}

%% file: macro&setup/myMacro.tex
\newtheorem{assumption}{Assumption}

\newcommand{\Obs}{y}

\newcommand{\Para}{\boldsymbol{\Theta}}
\newcommand{\TruePara}{\Para^{*}}

\newcommand{\rowdim}{d_1}
\newcommand{\coldim}{d_2}

\newcommand{\PolIdx}{\pi}

\newcommand{\TrInnProd}[2]{{\langle\kern-0.4ex\langle #1, #2
    \rangle\kern-0.4ex\rangle}}

\newcommand{\vertiii}[1]{{\left\vert\kern-0.25ex\left\vert\kern-0.25ex\left\vert #1 
    \right\vert\kern-0.25ex\right\vert\kern-0.25ex\right\vert}}

\newcommand{\Signal}[1]{\TruePara_{\PolIdx_{#1}}}

\definecolor{contcol1}{HTML}{72E094}
\definecolor{contcol2}{HTML}{24E2D6}
\definecolor{convcol1}{HTML}{C0392B}
\definecolor{convcol2}{HTML}{8E44AD}
